# Guidelines for evaluation of complex multi agent test scenarios

***UMGC: Project Research & Development of Methods for a Holistic Evaluation of Autonomous Vehicle Safety under Tropical Urban Conditions***

**CETRAN Team:**

Ana Isabel Garcia Guerra

Teng Sung Shiuan

(31 October 2023)

# TABLE OF CONTENTS

**Disclaimer**

**This research in this report is supported by the National Research Foundation, Singapore, and Land Transport Authority under Urban Mobility Grand Challenge (UMGC-L010).**

# 1. Motivation

Stakeholders of Autonomous Vehicles (AV) are facing a major scientific and technological challenge to demonstrate the safety of the autonomous vehicle in a complex environment and the vast number of potential driving situations the vehicle may encounter. The use of machine learning on AV systems containing non-explainable data sources for perception and knowledge management makes the formal verification (model based) difficult [1]. Empirical validation through testing techniques based on the analysis of the observed AV behavior in response to external stimuli seems to be the way due to the inability to apply model-based verification techniques [1].

In Singapore, the stage gate Milestone assessment framework for AVs before beginning trials on public roads utilize scenario-based testing concept as a form of empirical validation. For example, within the Milestone 3 assessment to enable trials without a safety driver, a framework has been developed for the generation and selection of relevant tests for scenario-based testing based on the desired operating design domain submitted by the applicant. This framework developed by the Centre of Excellence for Testing & Research of Autonomous Vehicle (CETRAN) supported by the Land Transport Authority (LTA), is today based on largely simplistic scenarios with defined parameters with mostly either one or no agents, and only occasionally more than one agent. Testing for edge cases is done by setting parameters for such a fundamental scenario to extreme values. However, many serious accidents and near-misses are edges cases which are not the result of extreme parameters, but of scenarios involving multiple agents. sequences of events and complex elements for AV.

This work aims to understand the source of complexity for AVs from traffic hazard breaking down the difficulties on AV capabilities as perception, situation awareness and decision-making. Guidelines for including complexity elements in test scenarios will be created from this understanding. This allows for clear structured manner of evaluating complexity elements incorporated in test scenarios and facilitates possible testing robustness of AV capability beyond edge cases.

Guidelines to evaluate complexity of multi agent test scenarios are composed by a list of elements to be considered in the future as selection criteria to generate or evaluate complexity of scenarios in any step of milestone assessment (e.g., static environment, actors, traffic management zone). The guidelines have been created in a structured way introducing multiple agents and scenario conditions that can be used to test the risk management ability of autonomous vehicles in a scenario-based test approach or complex traffic situations faced on road trials. One complexity element introduced in a test scenario could also create a complex scenario for assessment of an AV, without requiring multiple complexity elements to be present.

Guidelines are intended to be Autonomous Driving Systems (ADS) agnostic, considering autonomous vehicles with Society of Automotive Engineers (SAE) Level 4 & Level 5 levels [2] of autonomy with common actors (e.g., pedestrian, cyclist, vehicle, object), and common road infrastructure. These guidelines could be used when creating scenarios for any specified Operational Design Domain (ODD) based on the applicability of the elements. In addition, a selection of multi agent accident and incidents from a database of Singapore on road will be analysed using the above guidelines, to demonstrate the possible complexity elements for an AV if faced with such real on-road traffic situations.

# 2. Literature Review

Guidelines to evaluate complexity of multi agent test scenarios are structured in two main components, where the first component is the identification of hazards elements that could generate complexity leading to possible harm (incident or accident) when present in the AVs scenarios and the second component is the understanding of the source of complexity from each traffic hazards identified.

Literature review has focused on these two specifics topics, providing the intellectual foundation and context for conducting our own approach. The following sources presented have been relevant to guide the approach developed in this work.

## 2.1. SOTIF - ISO/PAS 21448

International Standards Organisation (ISO)/Publicly Available Specification (PAS) 21448 called Safety Of The Intended Functionality (SOTIF) standard [3] addressing vulnerabilities that are out of the scope of the Functional Safety Standard (FuSa) ISO 26262 [4]. The subject of both safety standards is the protection of humans from harm and injuries but while the objective of FuSa is to avoid unreasonable risks derived from hazards caused by malfunctioning behaviour of electrical and electronic components inside the vehicle, SOTIF´s objective is to avoid, control, and mitigate safety hazards that could happen on the environment of the vehicle without a system failure taking place. SOTIF applies to Artificial Intelligence (AI) and Machine learning systems, which rely on sensors and huge volumes of data to fed complex algorithms. FuSa and SOTIF are distinct and complementary aspects of safety that should be considered from the design to the validation phase of the AVs.

For this work, the SOTIF standard has been considered as a main source to understand complexity for an AV. SOTIF refers to the ability to the system to correctly comprehend the real-world environment (proper situation awareness), make right decisions and operate safely by ensuring the AV systems are operating within their design boundaries.

This includes the areas of:

- The situational awareness derived from complex sensors and processing algorithms.
- System robustness with respect to sensor input variation or diverse environment conditions

SOTIF addresses the challenges of ensuring the safety of intended functionality in complex and evolving real-world scenarios. SOTIF standard has been helpful to identify the system weaknesses and the related hazards that could trigger potentially unsafe behaviour of the system. Additionally, a list of example factors that could be considered in scenario creation is presented in Annex F, Table F.1 [3]. The SOTIF standard has been an inspiration for this work in traffic agent hazards to be considered, and understanding the complexity of these hazards for an AV.

## 2.2. Waymo – Building a credible Case for Safety

Waymo published a white paper [5] which applies SOTIF approach as a development stage to create a systematic framework to define their safety case. Their definition of safety case is based on the absence of unreasonable risk supported by the identification of scenarios containing triggering conditions in which the hazards can lead to harm and the application of risk assessment process. Waymo's white paper have been very useful for the identification and evaluation of agents hazards elements that could create complexity for AVs. These elements are considered triggering conditions for complex scenarios to evaluate the system capability.

## 2.3. Safety Demonstration of Automated Road Transport Systems

Following the publication of a national strategy for the development of autonomous vehicles in 2018, France has published a decree n° 2021-873 in 2021 which sets conditions for the deployment of automated vehicles and automated road transport systems on French roads [6]. It covers various levels of automation up to so-called "fully automated" systems provided that are under the supervision of a person in charge of remote intervention and are deployed on predefined routes or zones. This framework sets definitions and general safety provisions for these systems, as well as the liability regime and conditions of use for driver delegation systems in vehicles and automated road passenger transport systems

The decree also sets the conditions under which fully automated systems can be commissioned, following a specific safety demonstration process. This framework has been updated and adopted from September 1, 2022, allowing the deployment of automated passenger transport services, beyond an experimental framework [6]

The full system is validated by decision of the service organiser, after safety demonstration and opinion of an approved qualified body. Remote operators are able to intervene according to the system's conditions of use.

Various collective work was mobilized in the second half of 2022 intended to support stakeholders (system designers, operators, service organisers, and approved bodies) in the implementation of safety demonstrations. Reference documents (methodological documents or guides) were created to support this initiative. *Safety Demonstration of automated road transport Systems* [7] is one of these deliverables created within the framework of the working group on scenarios co-directed by the French Directorate General for Infrastructure, Transport and Mobility (DGITM)[1] and the SystemX Technological Research Institute (called the IRT SystemX)[2], with the participation of different stakeholders.

This is a methodological document dealing with generating, and enriching driving scenarios that can be used to demonstrate the safety of the system based on a structure of descriptors that can help in the search for completeness of testing and to identify the relevant complex scenarios challenging the AV system. The document has been valuable in establishing the structure of the approach used for this work and understand the source of complexity for AVs from traffic hazards. The list of descriptors from different sources presented in the document constitutes a useful input to address the hazards elements considered in the approach. The identification of "Elements of traffic agent hazards" classification considered in the approach has been guided by the "layers" described in this document.

[1]The Directorate General for Infrastructure, Transport and Mobility (DGITM) is a French central administration under the authority of the Minister of Ecology

[2] CETRAN has been collaborating with IRT System X during 2018-2023 on the joint project ASV (Autonomous Driving Simulation & Validation)

## 2.4. Rigorous Modelling and Validation of Autonomous Driving Systems

During the International Symposium on the Verification of Autonomous Mobile Systems (VAMS), held in Paris on the 9-10 March 2023, multiple invited presentations and round table discussions were conducted on the concerns of the Verification of Autonomous Mobile Systems operating in large outdoor spaces. The overall aim of the seminar was to promote the transfer of knowledge and experience between those involved in the design, development and governance of autonomous mobile systems operating across land, air, and sea applications, particularly those considered to be safety critical systems. CETRAN was involved in the organization of the event with other partners (IEEE Robotics Automation Society, IRT System X, Renault Group, The University of Manchester). More information about the Seminar could be found in this link https://www.irt-systemx.fr/evenements/vams-is-23

The plenary session presentation was conducted by Joseph Sifakis, on *"Rigorous Modelling and Validation of Autonomous Driving Systems"* [8]. He is a computer scientist and researcher, laureate of the 2007 Turing[3] Award for his work on verification method of computer hardware and software properties. This presentation focused on how the behaviour of an autonomous vehicle could be understood as a combination of functions. An AV consists of functions for achieving situational awareness by creating a model of its environment (perception and reflection), functions for decision-making (goal management and planning) and function for knowledge repository learned from driving data. It also introduced different types of complexity issues for the AVs, such as complexity of perception, complexity of uncertainty due to situations involving imperfect or unknown information such as dynamic change, rare events, and complexity of decision such as diversity of goals and size of the space of solutions for planning. This presentation has been very useful to establishing the breakdown of complexity of Autonomous System used by the approach in section 3.2.1

## 2.5. Testing System Intelligence

"Testing System Intelligence" is an article wrote by Sifakis in 2023 [1]. The article shows several examples illustrating the difficulty for assessing intelligent behaviour of AI systems. It also highlights the differences on the knowledge characteristic and the process for achieving situation awareness and decision making for both Human and Machine AI based.

It is difficult for AVs to achieve the same level of human situational awareness because human thinking is based on common sense knowledge and can combine symbolic mental models with concrete sensory knowledge, on other hand AVs can learn complex relationships and produce knowledge from multidimensional data, while humans show very limited capabilities in this type of task.

The article evaluates the existing validations techniques for AI systems and their limitations. It shows the non-applicability of model-based verification to machine learning systems due to non-explainable AI for perception and knowledge management, testing techniques as scenario-based testing validation based on experimental seems to be the only viable approach for Sifakis. A "replacement test" approach is proposed in the paper with the aims to evaluate the AI system ability to successfully replace a human in a specific task and context. This is interesting to manage the expectations for the deployment of ADS as it could be used as argument to support the justification that autonomous systems could be at least as successful as a human avoiding traffic accidents.

[3] The Turing Award is recognized as the Nobel Prize of Computing [20]

The document on the different mechanisms on how the AI systems knowledge database is developed, helping CETRAN to gain a deeper understanding of the complexity of situation awareness and decision making used by AI systems.

# 3. Approach

## 3.1. Introduction of approach

CETRAN's approach on creating guidelines for the generation or evaluation of complex scenarios is developed in two parts. First by creating guidelines as a list of elements of traffic agent hazards to be used in the future as selection criteria to generate or evaluate complexity of scenarios. The second part is the analysis of selected accidents or incidents from on road traffic that includes multi agent scenarios for complexity, using the guidelines created. These scenarios are representative of the variety of complexity elements identified in the guidelines.

To create the guidelines for complex scenarios to AVs, an understanding of AV architecture is necessary to breakdown the cause of complexity within the system architecture. This is then used in combination with elements of traffic agent hazards to link the source of complexity from the environment and the corresponding impact on the ADS performance at different function levels. This is elaborated upon in the following sections 3.2 and 3.2.2 below.

The expectations for deploying AVs on public road is based on the promise that their behaviour would be better or at least comparable with qualified human drivers. In the evaluation of what could be complex for AVs, a fundamental appreciation of the difference on the intelligence skills between human and machine should be first established. This allows for understanding of the gaps between human and machine and the corresponding complexities that arise from these gaps.

## 3.2. Guidelines Creation

### 3.2.1. Understanding Complexity of AVs

An understanding of the architecture of AV systems is required to understand what type of scenarios are complex for AVs when operating on roads. The overall flow system architecture of an AV is shown in Figure 1 below, where the external environment is sensed by hardware sensors on the AV. This information is then sent for processing on the onboard system which can be split into situation awareness and decision making. This then leads to activating the actuators on the vehicle such a as the hardware controls of steering, brake or accelerator to move the vehicle accordingly.

An autonomous vehicle falls under the broad category of an autonomous agent, in which its behaviour is understood as a combination of functions. The two overall functionalities are achieving situational awareness and decision-making [9]. Within situational awareness, the autonomous vehicle needs to "create a model of its own environment" [9] through perception and reflection – which is termed situational analysis for the remainder of this report. Goal management and planning of its path are functionalities within the decision-making process of the autonomous system architecture. These 4 aspects are highlighted on the right of Figure 1 below in the yellow and blue boxes. A fifth function not within the figure and not in scope within this work but should be acknowledged is the "production and application of knowledge to compensate for uncertainty and incomplete knowledge of the environment" [1], this could be from an example where the autonomous vehicle has pre-stored knowledge of maps which should complement with inputs of the perception function of its environment [1] such as lane marking or speed limit changes.

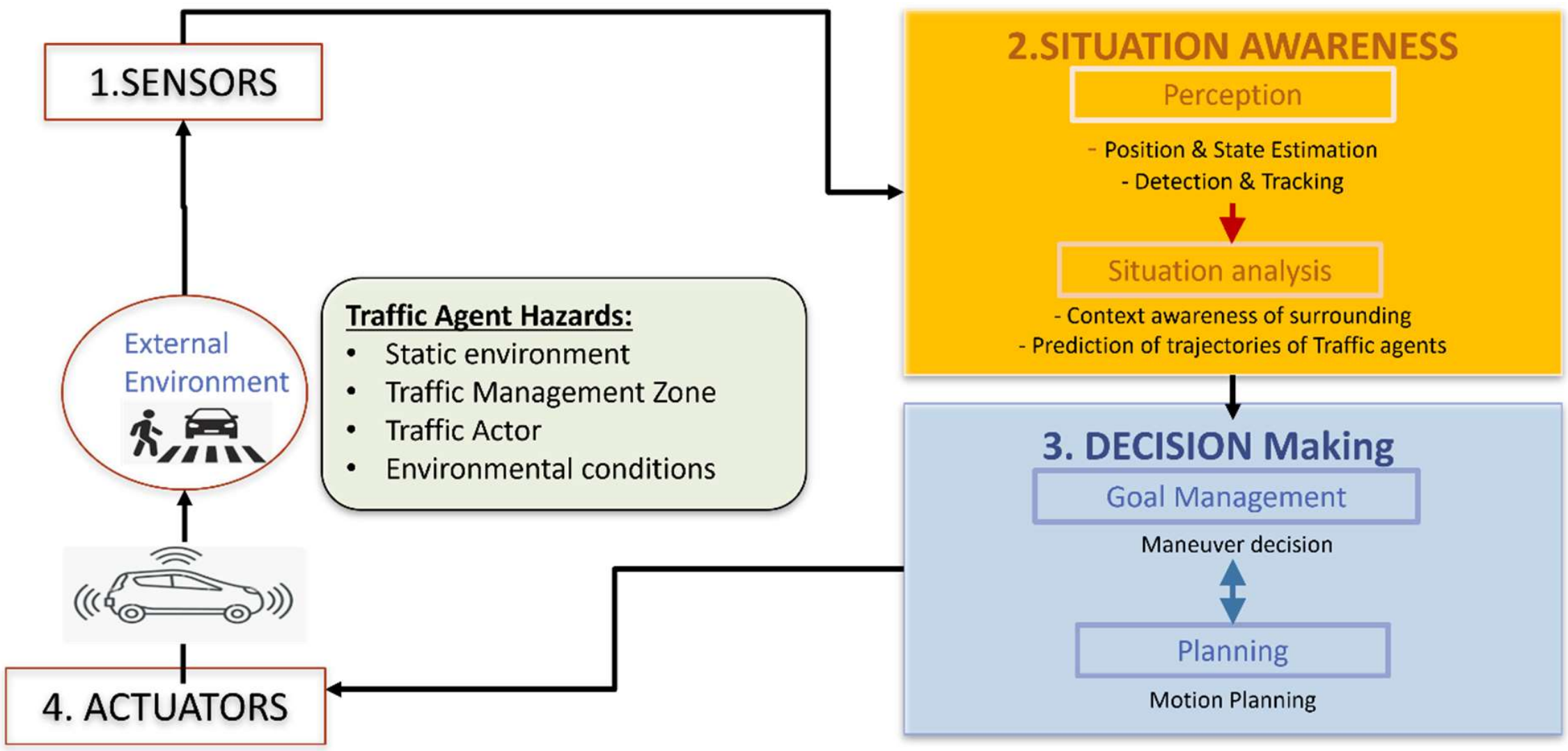

*Figure 1: Elements of AV architecture*

**Situation Awareness**

Within situation awareness, the first functionality of perception is focused on the AVs capability to interpret sensor data firstly on position and state estimation of the surrounding actors or ego, and secondly on the detection and tracking of actors. This is affected by both sensor capabilities and computing algorithm within the system. The second functionality of situation analysis consists of context awareness of the environment and the prediction of trajectories of other traffic agent hazards.

Situational awareness for humans is derived from common sense knowledge and combination of mental models with concrete sensory knowledge [1], machines such as AVs may not be able to match up the same levels of abstract reasoning as humans. One example would be understanding that multiple traffic cones placed on lane marking would mean the lane is closed, with a level of inferring as to whether the lane to the right or left of the cones is closed. This proved to be a difficult traffic situation to understand for a Waymo AV in 2021, where it attempted to drive between cones into the closed lane, later stopping between lanes and blocking traffic flow [10].

The functionality of perception is affected by the sensor capabilities used to detect and sense the external environment of surrounding traffic. The common choices of light detection and ranging (LiDAR), radar, vision camera each have their advantages and disadvantages. LiDAR has a wide range of view, with high range and angle resolution, while camera vision has availability of colour distribution and discernibility [11]. A disadvantage is that LiDAR's and vision camera's range of detection of 200 to 250 metres is greatly reduced under rain or fog conditions, compounded with high cost of LiDAR equipment and heavy calculation burden of camera vision [11]. This limits the maximizing of operation conditions on Singapore's tropical roads where rain is a common occurrence which could happen at any time of day. Radar, on the other hand, is applicable for all weather conditions but is of lower resolution compared to camera and LiDAR and is susceptible to noise and inapplicability for static object detection [11]. Relevant multi-sensor fusion strategy should be chosen that combines the advantages of each sensor such that the overall system runs cooperatively [11].

An overall example of situation awareness comprising perception and situation analysis (or reflection [1]) from typical traffic situation is on a wet road segment. A difficulty for an AV system could be that the perception by combining and analysing data from sensors does not detect the wet road condition, it is then unable to comprehend the situation (context awareness) and adapt to wet road condition that can lead to reduced adherence to the road surface due to water. For example, the vehicle may adjust its speed, apply brakes slowly or modify its trajectory to ensure safety.

**Decision- making**

The two functionalities within decision-making are first managing goals which would be strategy of the action to be taken, then planning where the action of carrying out the strategy is executed. Humans carry out the decision-making process through a value-based mechanism where they can "handle multiplicity of goals to satisfy our needs" with a choice based on a combination of value scale, somewhat a value balance depending on what is of priority [1]. This could be prioritising reachability of a route versus the safety to other road users amongst other dynamically changing goals while driving in traffic. This balance of goals and poorly understood value-based decision system could be a source of complexity of AVs to emulate and manage within the system algorithm. However, it could be said that machines could be taught to learn complex relationships and produce knowledge from multi-dimensional data, balancing goals quicker than a human if multiple goals were presented simultaneously. The complexity of decision-making increases as we move from a single goal to multiple goals and from a single agent to a system of agents (multi-agent).

### 3.2.2. Hazards Classification

With the foundation understanding of an AV system architecture and breakdown of where complexity can arise in the four functionalities of perception, situation analysis, goal management and planning, the next step would be to classify the external environment of traffic interactions. This is termed the traffic agent hazards in this paper, where the elements are in four parts such as static environment, traffic management zone, traffic actors and environmental conditions. The four parts were derived by CETRAN to split the elements of traffic agent hazards into the four broad parts when creating scenarios. The elements considered took reference to a publication by the French Ministry of Ecological Transition study of scenarios for automated vehicles [7], adapted for Singapore urban traffic.

1. Static environment:
    - This focuses on the permanent physical aspects of road infrastructure. It includes categories such as traffic signs, traffic lights and road markings. Semi-permanent road changes are considered within this section of Static Environment, under its respective changes to lane markings or traffic signs or road infrastructure.
2. Traffic Management Zone
    - This focuses on managing traffic flow, these could be linked to changes to the physical aspects of road infrastructure such temporary road lane closure and work-zones.
3. Traffic Actors
    - This focuses on the actors that could be present on road for traffic interactions. This includes actor behaviours or interactions and visibility of actors. Common actors are pedestrians, vehicles and objects.
4. Environmental Conditions
    - This focuses on the environmental conditions that could happen temporarily during a drive.

Breaking down external environment traffic agent hazards offers granularity in elements of on-road traffic scenarios. A specific traffic agent hazard from the external environment on road is then analysed as source of complexity to AVs in any of the four functions described earlier in Section 3.2 above. An example would be a non-functioning traffic light could cause complexity for an AV in any of the four functions described earlier in Section 3.2 above, or in combination. This general approach of the traffic agent hazards as a source of complexity and the corresponding breakdown into the four functions is shown pictorially in Figure 2 below.

**Traffic Agent Hazards:**

| Elements of Traffic Agent Hazards | Categories (of elements) |
|---|---|
| Static environment | 1. Traffic signs<br>2. Traffic lights<br>3. Road marking<br>4. … |
| Traffic Management Zone | 1. Temporary lane closure<br>2. Work-zone<br>3. … |
| Traffic Actor<br>*With common actors:*<br>*(e.g. pedestrian, cyclist, vehicle, object)* | 1. Actor behaviour/interaction<br>2. Visibility of actors |
| Environmental Conditions | 1. Road conditions<br>2. Lighting<br>3. .. |

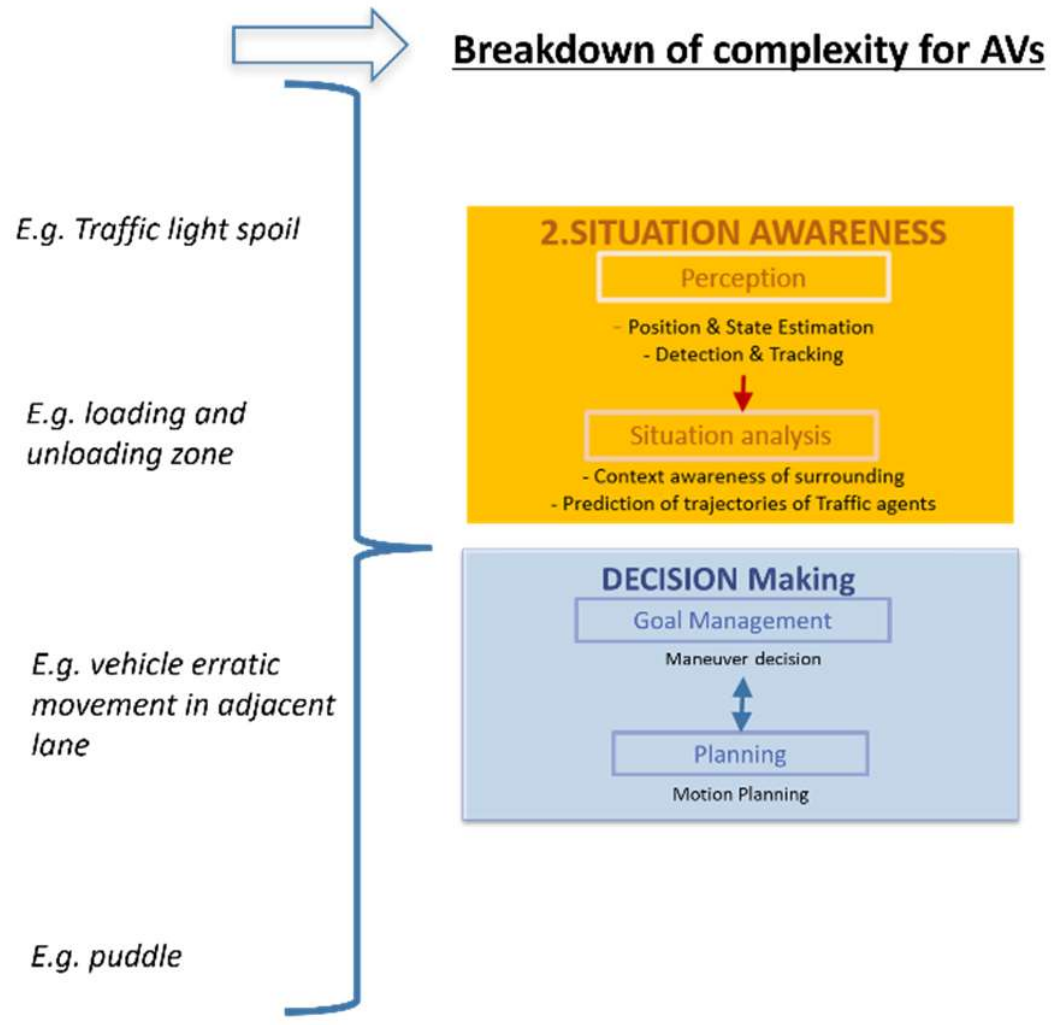

*Figure 2: Elements of Hazards in Traffic Environment*

## 3.3. Analysis of Identified Scenarios

In the definition of complex scenarios for AVs for assessment, it is desirable to create from elements of traffic agent hazards that causes complexity from real traffic situations.

It is necessary to obtain real traffic situations for the applicability of complexity in real traffic. CETRAN had purchased and obtained access of Resembler webtool that contains a database of at least 1000 real traffic accidents and incidents that are obtained from publicly available dashcam recordings. These accidents and incidents are a basis of real traffic situations that an AV would encounter on Singapore's urban traffic. Although accidents and incidents occur occasionally as a result of drivers not following rules of the road, it provides a glimpse into the possible difficult situation an AV would need to react to. One complexity of operating safely in Singapore roads is the dense traffic and driving in mixed traffic flow with other human drivers. Understanding some of these incidents provide real insights and is a step with increased realism from abstract scenario creation.

## 3.4. Collaboration with Resembler

A Research Collaboration Agreement (RCA) was established with Resembler. The Resembler team have experience in AV development and deep knowledge of the accidents and incidents database. The RCA allows for open discussions in getting feedback on AV difficulties in the selected accidents and incidents used for analysis. The Resembler team provided suggestions on certain examples that highlights the differences between how an AV and human could differ in reacting to a certain traffic situation. This helped CETRAN to down-select accidents and incidents for analysis of complexity that will be shown in Section 5, to cover the various complexity elements in the guidelines.

# 4. Guidelines of evaluation of scenarios

The guidelines for the evaluation of scenarios are listed within this section based on the four parts of traffic agent hazards and their corresponding complexity reason within the AV architecture. It is structured based on external traffic agent hazards such as the surrounding infrastructure and actors such as the interaction between an AV and actors in this environment on Singapore roads. This is further clarified in the classification structure in Section 4.1, followed by the detailed guidelines of complexity elements and their corresponding reason of complexity for AVs are listed in Section 4.2.

## 4.1. Classification Structure

The elements of traffic agent hazards were earlier mentioned to be classified into 4 parts; static environment, traffic management zone, traffic actor and environmental conditions. Within each part, there are further classification characteristics to organise these elements. This allows for more specific categorisation of elements, with some examples listed for inspiration that could be used for complex scenario generation or evaluation.

1. Static environment:
    - This focuses on the permanent physical aspects of road infrastructure. It includes categories such as traffic signs, traffic lights and road markings. Semi-permanent road changes are considered within this section of Static Environment, under its respective changes to lane markings or traffic signs or road infrastructure.[4]
2. Traffic Management Zone
    - This focuses on managing traffic flow, these could be linked to changes to the physical aspects of road infrastructure such temporary road lane closure and work-zones.
3. Traffic Actors
    - This focuses on the actors that could be present on road for traffic interactions. This includes actor behaviours or interactions and visibility of actors. Common actors are pedestrians, vehicles and objects.
4. Environmental Conditions
    - This focuses on the environmental conditions that could happen temporarily during a drive.

[4] One example is the North-South Corridor construction that leads to road layout changes around Thomson Road [12]. The road layout changes fall under Section 4.2.1.1.4.5 because the road infrastructure has been changed, such as traffic intersection and lane markings. But the environment there also has elements of work-zone which is used from section 4.2.2.

## Static Environment

### Physical Infrastructure
- Visibility
- Road conditions
- Interference zones
- Road Infrastructure/type
  - Lanes
  - Signalized junctions
  - Unsignalized junctions
  - Special zones
  - Changes to road infrastructure
- Road marking legibility *(readability)*
- Others

### Traffic Signs
- Legibility *(readability)*
- Visibility of sign
- Changes to Traffic Signs

### Traffic Lights
- Legibility *(readability)*
- Visibility of lights
- Changes to Traffic Lights

## Traffic Management Zone

- Work-zone
- Other
- Human directed traffic
- Visibility of zone information

## Traffic Actor

### Vehicles
- Vehicle Lights
  - Intention indicators
  - Emergency vehicle lights
- Types
  - Nominal
  - Special
- Static State
  - Nominal
  - Behaviour against traffic rules
  - Visibility
- Dynamic State
  - Nominal
  - Erratic
  - Behaviour against traffic rules
  - Visibility

### Pedestrian
- Type
  - With accessories
  - Nominal
- Static State
  - Nominal
  - Behaviour against traffic rules
  - Visibility
- Dynamic State
  - Nominal
  - Behaviour against traffic rules
  - Erratic
  - Visibility

### Object
(near/ on road surface)
- Static State
  - Permanent
  - Temporary
  - Visibility
- Dynamic State
  - Temporary
  - Visibility

## Environmental Conditions

### Visibility
- Weather
- Suspended particles
- Lighting

### Others
- Weather induced road conditions
- Weather elements

## 4.2. Table of elements and breakdown

In this following section, details of elements and its corresponding breakdown of complexity reasons for an AV are listed based on the four AV functions within its system architecture are tabulated. Each element is considered to have one main impact of complexity on the AV architecture, with the possible cascading effect also listed when applicable.

### 4.2.1. Static Environment

Static environment focuses on the permanent physical aspects of road infrastructure. It includes categories such as traffic signs, traffic lights and road markings. Semi-permanent road changes are considered within this section of Static Environment, under its respective changes to lane markings or traffic signs or road infrastructure.

#### *4.2.1.1. Physical Infrastructure*

##### *4.2.1.1.1. Road Visibility*

This refers to different types of hazards occluding the view of road for the AV.

| **Example Category** | **Examples** | **Perception** | | **Situation Analysis** | | **Goal Management** | **Planning** |
|---|---|---|---|---|---|---|---|
| | | *-Position & State Estimation* | *-Detection & Tracking* | *-Context Awareness* | *-Prediction* | | |
| Other infrastructure or hazards occluding the road | At U-turn: plant blocking view of road<br>Sidewall/ noise-barrier occluding the view of road; | *No information of position and state of other road users that are located in the occluded view of the road.* | *AV may be unable to detect road users that are located in the occluded view of road, which could result in AV not having information for tracking other road users* | *If the AV is not able to perceive the complete road it will not be able to understand the driving context and to react accordingly to the situation* | *If the AV is not able to perceive the complete road it will not be able to predict the future movement of the road agents inside and to react accordingly to the situation* | *If the AV is not able to perceive the complete road it could lead to difficulty in balancing of goals between traffic flow, safety distance to other actors, reachability. E.g., It could decide to perform the U turn when an oncoming vehicle is on the road* | *Road visibility affects knowledge of available space to carry out manoeuvre.* |

### *4.2.1.1.2. Road Conditions*

This section focuses on the factors on the road that could affect the drive.

| **Example Category** | **Examples** | **Perception** | | **Situation Analysis** | | **Goal Management** | **Planning** |
|---|---|---|---|---|---|---|---|
| | | *-Position & State Estimation* | *-Detection & Tracking* | *-Context Awareness* | *-Prediction* | | |
| Surface material reflectivity | Asphalt/ cement | *Highly reflective road materials (E.g., Freshly paved asphalt surfaces often have a smooth and glossy texture. could cause noise in the sensor data. This could influence the accuracy of object position estimation.* [12] | *Reflectivity levels of different road surfaces could have an impact on how the light is reflected to the sensors (Lidar, camara) The accuracy of object detection of the different actors could be reduced.* | *AV may not have awareness to understand the ego's capability for these road surfaces (detection accuracy limitation, level of adherence...) to adapt its behaviour accordingly* | *If the AV is not able to perceive and analyze these conditions, it will not be able to predict accurately the future movement of the actors in the environment and to react accordingly to the situation.* | *Could be difficult to achieve AV's reachability versus safety. E.g., Vehicle's speed/braking strategy for tackling pothole and relevant size.* | *Road condition (Potholes/reflectivity) could affect space where manoeuvres are to be carried out in.* |
| Surface deterioration | Pothole | | *Road condition (Potholes) could be difficult to detect and classify.* | *AV awareness to understand the ego's capability to tackle pothole and other actors' possible maneuver to avoid.* | | | |

### *4.2.1.1.3. Interference Zones*

This focuses to factors that could affect the AV sensors, limiting the performance.

| **Example Category** | **Examples** | **Perception** | | **Situation Analysis** | | **Goal Management** | **Planning** |
|---|---|---|---|---|---|---|---|
| | | *-Position & State Estimation* | *-Detection & Tracking* | *-Context Awareness* | *-Prediction* | | |
| On-road | Tall buildings/ tunnel/ tree-cover | *Infrastructure affect Ego's localization accuracy and can lead to performance limitation on actor's position measurement.* | | *Inaccuracy of localization (ego leading to actor estimation) could lead to wrong understanding of surrounding actor interactions and intentions* | | *Inaccuracy of localization (ego leading to actor estimation) could lead to difficulty in balancing safety distances to actors versus AV right of way.* | *Localization accuracy could affect accuracy in space available for manoeuvres, subsequent manoeuvre could infringe rules.* |
| Near-road infrastructure | Multi-story carpark/ underground carpark/ | | | | | | |

### *4.2.1.1.4. Road infrastructure /type*

This section is to highlight the potential complexity based on road infrastructure or road zones, further interactions with other traffic actors are listed in the following section 4.2.3 (Traffic Actors)

#### *4.2.1.1.4.1. <u>Lanes</u>*

| **Example Category** | **Examples** | **Perception** | | **Situation Analysis** | | **Goal Management** | **Planning** |
|---|---|---|---|---|---|---|---|
| | | *-Position & Estimation* | *-Detection & Tracking* | *-Context Awareness* | *-Prediction* | | |
| Lane variation | 1. Merging<br>2. Splitting lanes | | | *AV may have difficulty understanding the pattern of traffic flow.* | *AV may have difficulty predicting the path of actors'* | *AV may have difficulty strategizing its position relative to traffic in such situation if it did not have context* | *AV may have difficulty in carrying out the execution manoeuvre into* |
| | | | | *E.g., Alternate slotting in of vehicles (left and right).*<br>*E.g., lane direction changes of traffic ahead, or cars slowing down ahead for speed bumps* | *actions at these lanes.* | *awareness, merging lane (to slow down and slot in between vehicles) or in splitting lanes to keep left/right into desired lane.* | *traffic maintaining safety distances to other vehicles* |
| Lane discipline | 1. Bus lane<br>2. Lane direction change ( straight to right turn only)<br>3. Yellow-box<br>4. Hump | | | | | | |

#### 4.2.1.1.4.2. ***Traffic Junctions - Signalized***

| **Example Category** | **Examples** | **Perception** | | **Situation Analysis** | | **Goal Management** | **Planning** |
|---|---|---|---|---|---|---|---|
| | | *-Position & Estimation* | *-Detection & Tracking* | *-Context Awareness* | *-Prediction* | | |
| Right-of-way indication | *T-junction/ cross- junction / signalized pedestrian crossing* | | | *AV may have difficulty understanding the pattern of traffic flow of other actors at intersection. It may not realise if others may be infringing rules of the road.*<br>*e.g., Does not understand the potential for cars waiting to carry out discretionary right turn* | *AV may have difficulty predicting the path of actors' actions at these junctions.*<br>*e.g., actor potentially carrying out discretionary right turn from stopped position.* | *AV may have difficulty in maintaining right of way versus safety distances to other vehicle.* | *AV may have difficulty in carrying out the execution manoeuvre into traffic maintaining safety distances to other vehicles.* |
| Non right-of-way indication | Discretionary right turn | | | | | | |

#### 4.2.1.1.4.3. ***Traffic Junctions – Unsignalized***

<table>
<tr><th rowspan="2">Example category</th><th rowspan="2">Examples</th><th colspan="2">Perception</th><th colspan="2">Situation analysis</th><th rowspan="2">Goal management</th><th rowspan="2">Planning</th></tr>
<tr><th>-Position & estimation</th><th>-Detection & tracking</th><th>-Context awareness</th><th>-Prediction</th></tr>
<tr><td>Minor-major Road intersections</td><td>U turn/ left turn slip road / car park entry/ minor roads</td><td></td><td></td><td rowspan="3">AV may have difficulty understanding the pattern of traffic flow and priorities of other actors at intersection.<br>E.g., At pedestrian crossings, understanding of typical pedestrian behaviour and priority.</td><td rowspan="3">AV may have difficulty predicting the path of actors at these junctions.<br>e.g., if other actors have right of way to U-turn and not just turning right.<br>e.g., if actors entering or exiting roundabout.<br>e.g., if pedestrians are intending to cross.</td><td rowspan="3">AV may have difficulty in maintaining goals of traffic flow versus safety distances to other vehicles/ pedestrians.<br>e.g., when to enter roundabout.</td><td rowspan="3">AV may have difficulty in carrying out the execution manoeuvre into traffic maintaining safety distances to other actors.</td></tr>
<tr><td>Pedestrian crossing</td><td>Zebra-crossing & non-zebra crossing</td><td></td><td></td></tr>
<tr><td>Other</td><td>Roundabout</td><td></td><td></td></tr>
</table>

#### 4.2.1.1.4.4. ***Special Zones***

<table>
<tr><th rowspan="2">Example Category</th><th rowspan="2">Examples</th><th colspan="2">Perception</th><th colspan="2">Situation Analysis</th><th rowspan="2">Goal Management</th><th rowspan="2">Planning</th></tr>
<tr><th>-Position & Estimation</th><th>-Detection & Tracking</th><th>-Context Awareness</th><th>-Prediction</th></tr>
<tr><td>Vulnerable Road User (VRU) Zone</td><td>1. School zone<br>2. Silver zone<br>3. Bus-stop/ Bus pick up zone</td><td></td><td></td><td>AV may have difficulty understanding the pattern of traffic flow of other actors approaching the special zone. It also signifies a</td><td>AV may have difficulty predicting the path of actors at these zones</td><td>Difficulty in maintaining goals of traffic flow versus safety distances to other actors. And possible erratic VRU actions.</td><td>AV may have difficulty in carrying out the execution manoeuvre maintaining</td></tr>
</table>

| | | | | | | | |
|---|---|---|---|---|---|---|---|
| | | | | *possible concentration of VRU here.<br>E.g., Cars could be stopping often to drop off/pick up school goers, stopping time is short. Awareness of door opening.<br>E.g., Bus could be stopping often to drop off/pick up passengers. Pedestrians could cross ahead of bus (occluded)* | | *E.g., Should AV change lane before approaching school zone - might be blocking drop off/ pick up lane if keeping left.<br>E.g., Should AV change lane before approaching stopped bus.* | *safety distances to other actors* |

#### 4.2.1.1.4.5. ***Changes to road infrastructure/ type***

| **Example category** | **Examples** | **Perception** | | **Situation analysis** | | **Goal management** | **Planning** |
|---|---|---|---|---|---|---|---|
| | | *-Position & estimation* | *-Detection & tracking* | *-Context awareness* | *-Prediction* | | |
| Change of Physical infrastructure | -Zones: New school zone/ New silver zone.<br>-Unsignalized junction: new pedestrian crossing<br><br>-Lanes: Change of road lane direction/lane position | | | *If AV uses HD maps only, this information may not be updated. Otherwise, if AV uses mix of HD maps + cameras to detect change of physical infrastructure, this could lead to difficulty in reconciling information between HD map and sensor, such as drivable area and traffic movement.* | *If AV does not update its HD maps or reconcile updated physical infrastructure with cameras, it will not be able to predict the behaviour of the other road users that can understand change in static environment.* | *AV may have difficulty in maintaining goals of traffic flow versus safety distances to other vehicles/ pedestrians* | *Change in physical infrastructure affect AV's understanding of driving area, could affect space where manoeuvres are to be carried out in.* |

#### *4.2.1.1.5. Road Markings Legibility (readability)*

This section refers to the legibility or readability of the road marking.

| **Example category** | **Examples** | **Perception** | | **Situation analysis** | | **Goal management** | **Planning** |
|---|---|---|---|---|---|---|---|
| | | *-Position & estimation* | *-Detection & tracking* | *-Context awareness* | *-Prediction* | | |
| Deterioration | 1. Lane Marking<br>2. Erased pedestrian crossing | | *Markings could be difficult to detect.* | *If AV uses mix of HD maps and cameras to detect lane markings this could lead to difficulty in reconciling information between HD map and sensor:*<br>*1. Understanding drivable area and understanding traffic flow.*<br>*2. Understanding special road zones such as pedestrian crossing*<br>*If AV uses only HD maps to detect road markings, it will not be able to understand the behavior of the other road users that could perceive the deterioration of the road markings and change its behaviour* | *If the AV has difficulties to understand the context it will not be able to predict with accuracy the behaviour of the other road users that can understand change in static environment.* | *Difficult for AV to manage its goal is to be in middle of drivable area if detection of lane marking affects definition of drivable area* | *Deteriorated lane marking affect AV's understanding of drivable area and could affect space where manoeuvres are to be carried out in.* |

#### 4.2.1.1.6. *Others*

| Example category | Examples | Perception | | Situation analysis | | Goal management | Planning |
|---|---|---|---|---|---|---|---|
| | | *-Position & estimation* | *-Detection & tracking* | *-Context awareness* | *-Prediction* | | |
| Barriers | Carpark barriers | | *AV might not be able to classify the barrier correctly* | *AV may have difficulty understanding the movement of the barrier and its context of allowing a vehicle to enter or not.* | | *AV may have difficulty in maintaining goals of reachability versus safety to objects.* | *AV may have difficulty in judging space available where manoeuvres are to be carried out in.* |

### 4.2.1.2. *Traffic Signs*

#### 4.2.1.2.1. *Legibility of signs (readability)*

| Example Category | Examples | Perception | | Situation Analysis | | Goal Management | Planning |
|---|---|---|---|---|---|---|---|
| | | *-Position & State Estimation* | *-Detection & Tracking* | *-Context Awareness* | *-Prediction* | | |
| Sign deterioration | Erased, damaged, twisted.<br><br>Fallen into the ground | | *If AV uses mix of HD maps and cameras to detect signs the loss of legibility of the sign could lead to difficulty to detect, track, and classify the sign* | *If AV uses mix of HD maps and cameras to detect signs, the loss of legibility of the sign could lead to difficulty to understand the situation and to define the adapted AV behaviour.*<br>*If AV uses only HD maps to detect signs, it will not be able to understand the* | *If the AV has difficulties to understand the context it will not be able to predict with accuracy the behaviour of the other road users that can understand change in static environment.* | *If AV have difficulty in perceiving information from the traffic signs, its could affect the balancing of goals between traffic flow and safety to other actors.* | *AV could have difficulty in using information from the traffic signs, its manoeuvre could conflict with traffic sign information.* |
| | | | | *behavior of the other road users that are affected by the deterioration of the signs.* | | | |

### *4.2.1.2.2. Visibility of signs*

This section focuses on the occlusion of the traffic signs, this could result from other traffic hazards such as static environment and traffic actors.

| Example Category | **Examples** | **Perception** | | **Situation Analysis** | | **Goal Management** | **Planning** |
|---|---|---|---|---|---|---|---|
| | | *-Position & State Estimation* | *-Detection & Tracking* | *-Context Awareness* | *-Prediction* | | |
| Infrastructure occluding the traffic signs | Wall/Other/Scaffolding occluding Traffic sign | | *If AV uses mix of HD maps and cameras to detect signs and the sign is occluded in a fugitive or permanent manner, this could lead to a difficulty detecting, tracking, and classifying the sign* | *If AV uses mix of HD maps + cameras to detect signs, and the sign is not visible permanent or temporary, it could lead to difficulty to understand the situation and to define the adapted AV behaviour .*<br>*If AV uses only HD maps to detect signs, it will not be able to understand the behaviour of the other road users that could perceive the sign and change its behaviour.* | *If the AV has difficulties to understand the context it will not be able to predict with accuracy the behaviour of the other road users that can understand change in static environment.* | *If AV has difficulty perceiving information from the traffic signs, it could affect in its balancing of goals between traffic flow and safety to other actors.* | *AV could have difficulty in using information from the traffic signs, its manoeuvre could conflict with traffic sign information.* |
| Temporary hazards occluding the traffic signs | Vegetation occluding traffic sign | | | | | | |
| | Work-zone blocking traffic sign | | | | | | |
| | Vehicle blocking traffic sign | | | | | | |

#### 4.2.1.2.3. *Changes to traffic signs*

| Example Category | Examples | Perception | | Situation Analysis | | Goal Management | Planning |
|---|---|---|---|---|---|---|---|
| | | *-Position & State Estimation* | *-Detection & Tracking* | *-Context Awareness* | *-Prediction* | | |
| Change of traffic signs | -Signboard information updates<br><br>-Location of traffic signs moved | | | *If AV uses HD maps only, this information may not be updated and have difficulty to adapts its own behaviour to the new sign.<br>Otherwise, if AV uses mix of HD maps + cameras to detect traffic signs, this could lead to difficulty in reconciling information between HD map and sensor, such as drivable area and traffic movement.* | *If AV has difficulty in context awareness, it will not be able to predict the behaviour of the other road users that can understand change in static environment.* | *If AV has difficulty in context awareness it could affect the balancing of goals between traffic flow and safety to other actors.* | *AV could have difficulty in context awareness, its manoeuvre could conflict with traffic sign information* |

### 4.2.1.3. *Traffic Lights*

#### 4.2.1.3.1. *Type*

| Example Category | Examples | Perception | | Situation Analysis | | Goal Management | Planning |
|---|---|---|---|---|---|---|---|
| | | *-Position & State Estimation* | *-Detection & Tracking* | *-Context Awareness* | *-Prediction* | | |
| Multiple traffic lights in close proximity | Junction with multiple lights next to each other for 1. left turn slip road and 2. lane going straight. | *In this kind of situation, AV needs to ensure accuracy of position of the lights to guarantee the identification of the relevant light* | *AV might have difficulty in classification of the lights, either identify as a group or separate detections.* | *Difficulty to understand the traffic light intended for its path and to define the adapted AV behaviour.* | *AV could have difficulty in prediction of traffic flow due to wrong classification and context awareness.* | *If AV has difficulty perceiving information from the traffic lights it could affect the balancing of goals between traffic flow and safety to other actors.* | *AV could have difficulty in understand the traffic light intended for its path, its manoeuvre could conflict with traffic light information.* |

#### 4.2.1.3.2. *Legibility (readability)*

| Example Category | Examples | Perception | | Situation Analysis | | Goal Management | Planning |
|---|---|---|---|---|---|---|---|
| | | *-Position & State Estimation* | *-Detection & Tracking* | *-Context Awareness* | *-Prediction* | | |
| Traffic light deterioration | 1. No lights - Out of order (no light) or fallen onto ground<br><br>2. Limited Lighting-Blinking strangely/ half | | *The loss of legibility of the light could lead to difficulty to detect, classify the colour / state of the light* | *AV might have difficulty to understand the situation and to define the adapted AV behaviour.*<br>*If the deterioration is detected by the AV, the AV would need to get* | *AV could have difficulty in prediction of traffic flow due to different interpretation of traffic light deterioration.* | *If AV have difficulty in perceiving information from the state of the lights it could affect the balancing of goals and the safety decision. AV would* | *AVs could have difficulty in planned manoeuvre adapted to the traffic flow.* |
| | lighted / pole twisted | | | *awareness of the context based on the status of other lights on the surrounding area. Or based on other data such actor's movement detection.* | | *need to consider the confidence on the alternative information used for situation awareness and adapt towards a conservative behaviour for managing traffic flow.* | |

### *4.2.1.3.3. Visibility of Traffic Lights*

This section focuses on the occlusion of the traffic lights, this could result from other traffic hazards such as static environment and traffic actors.

| **Example Category** | **Examples** | **Perception** | | **Situation Analysis** | | **Goal Management** | **Planning** |
|---|---|---|---|---|---|---|---|
| | | *-Position & State Estimation* | *-Detection & Tracking* | *-Context Awareness* | *-Prediction* | | |
| Infrastructure occluding the traffic lights | Wall/Other/Scaffolding occluding Traffic lights | | *Difficulty to detect and classify light information if the light is occluded partially or completely.* | *AV might have difficulty to understand the situation and to define the adapted AV behaviour. AV would need to get awareness of* | *Limitation of the perception can lead to incorrect prediction of traffic flow.* | *If AV have difficulty in perceiving information from the state of the lights, it could affect the balancing of* | *AVs could have difficulty in planned manoeuvre adapted to the traffic flow.* |
| Temporary hazards occluding the traffic lights | Vegetation occluding traffic lights | | | | | | |
| | Work-zone blocking traffic lights | | | | | | |
| | Vehicle blocking traffic lights | | | *the context based on the status of other lights in the surrounding area or based on other data such actor's movement detection.* | | *goals and the safety decision. AV would need to consider the confidence on the alternative information used for situation awareness and adapt towards a conservative behaviour for managing traffic flow.* | |

#### 4.2.1.3.4. *Changes to Traffic Lights*

| **Example category** | **Examples** | **Perception** | | **Situation analysis** | | **Goal management** | **Planning** |
|---|---|---|---|---|---|---|---|
| | | *-Position & estimation* | *-Detection & tracking* | *-Context awareness* | *-Prediction* | | |
| Change of traffic lights | Change in traffic light positions.<br><br>Change in traffic lights operation: (discretionary right turn to controlled right turn) | | | *This could lead to difficulty in reconciling information between HD map and sensor, such as drivable area and traffic movement.* | *If AV does not update its HD maps or reconcile updated traffic lights with cameras, it could have difficulty to predict the behaviour of the other road users that can understand the change.* | *If AV has difficulty perceiving information from the traffic signs, it could affect the balancing of goals between traffic flow and safety to other actors.* | *Change in traffic lights affect AV's understanding of driving area, could affect space where manoeuvres are to be carried out and adapting to traffic flow* |

## 4.2.2. Traffic Management Zone

Traffic management zone focuses on managing traffic flow, these could be linked to changes to the physical aspects of road infrastructure such temporary road lane closure and work-zones.

### *4.2.2.1. Work Zone*

This section focuses on the elements that are on the road or on the roadside related to work-zone or construction areas and could sometimes lead to lane closure.

| **Example Category** | **Examples** | **Perception** | | **Situation Analysis** | | **Goal Management** | **Planning** |
|---|---|---|---|---|---|---|---|
| | | *-Position & State Estimation* | *-Detection & Tracking* | *-Context Awareness* | *-Prediction* | | |
| Work-zone elements | Cones/Barriers | | *Where work zone elements are present, the AV might have difficulty in correct classification. The performance of the classification will depend on the previous training data available. Difficulty is linked to the diversity of the configurations involved.* | *Difficulty in understanding the meaning of the signboards/ work zone elements (e.g. cones) /information (e.g. EMAS) associated to the zones and define a good behaviour for each situation encountered (e.g., slow down, stop, works ahead.)* | *If AV has difficulty in understanding work-zone elements, it will not be able to predict the behaviour of the other road users who are able to perceive correctly the sign and could have a different behaviour than expected.* | *Could be difficult to achieve AV's target if lane is narrowed from work-zone elements (changing lane /rerouting).* | *AV could have difficulty in accurate estimation space available for its planned manoeuvres.* |
| | Temporary signboard | | | | | | |
| | Dynamic traffic signs (could be Expressway Monitoring and Advisory System (EMAS)) | | | | | | |
| | Stop/Go Boards | | | | | | |

#### 4.2.2.2. Other

This section focuses on the elements that are on the road or on the roadside related to other elements in traffic management zone such as vehicle breakdown.

| Example Category | Examples | Perception | | Situation Analysis | | Goal Management | Planning |
|---|---|---|---|---|---|---|---|
| | | *-Position & State Estimation* | *-Detection & Tracking* | *-Context Awareness* | *-Prediction* | | |
| Other Elements | Triangle to signal car breakdown | | *Classification of breakdown triangle not known - random object. May not be detected if too reflective (camera)?* | *AV could have difficulty understanding situation- possibly keep distance from accident zone.* | *AV could have difficulty predict the behaviour of the other road users who are able to understand the situation.* | *Could be difficult to achieve AV's reachability if driving lane is affected by accident (changing lane /rerouting), versus maintaining safety distance to accident zone.* | *AV could have difficulty in accurate estimation space available for its planned manoeuvres.* |

#### 4.2.2.3. Human directed traffic

This section focuses on the component of traffic management zone that comes from *human intervention* that could be found at lane closures from loading and unloading zones or work-zones, and traffic light breakdown incidents.

| Example Category | Examples | Perception | | Situation Analysis | | Goal Management | Planning |
|---|---|---|---|---|---|---|---|
| | | *-Position & State Estimation* | *-Detection & Tracking* | *-Context Awareness* | *-Prediction* | | |
| Human directed traffic elements | Hand signalling | | *Hand signaling may not be classified separately or correctly (not pedestrian)* | *AV may not understand hand signal intentions for maintaining traffic flow (move ahead, stop, move to right/left, slow down)* | *AV could have difficulty predict the behaviour of the other road users who are able to understand the situation.* | *AV might have difficulty in balancing goals of reachability may not be achieved if hand signals don't allow to continue path and maintaining traffic flow* | *AV could have difficulty in accurate estimation space available for its planned manoeuvres.* |

#### 4.2.2.4. *Visibility of traffic zone information*

This section focuses on the occlusion of the traffic management zone elements, this could result from other traffic hazards such as static environment and traffic actors.

| **Example Category** | **Examples** | **Perception** | | **Situation Analysis** | | **Goal Management** | **Planning** |
|---|---|---|---|---|---|---|---|
| | | *-Position & State Estimation* | *-Detection & Tracking* | *-Context Awareness* | *-Prediction* | | |
| Other users or infrastructure occluding the traffic zone information | -Cones/ barriers/others temporary elements occluded by a parked car<br><br>-Signboards not visible due to road curvature | *AV is unable to have localisation information of occluded traffic management zone elements* | *Unable to detect traffic management zone elements. No classification/tracking available.* | *If AV can't perceive the management zone, it could lead to difficulty to understand the situation and to define the adapted AV behaviour. E.g. did not realise work-zone ahead which requires AV to slow down (and other traffic)* | *If AV can't perceive the management zone, it could lead to difficulty to predict the behaviour of the other road users who are able to understand the situation and could have a different behaviour than expected.* | *Visibility of traffic management zone affects goal management, If the AV is not able to see the work-zone ahead, it will not be able to decide on safe behaviour (e.g. slow down due to work zone ahead).* | *AV could have difficulty in accurate estimation space available for its planned manoeuvres.* |

### 4.2.3. Traffic Actors

This focuses on the actors that could be present on road for traffic interactions. This includes actor behaviours or interactions and visibility of actors. Common actors are pedestrians, vehicles and objects.

#### *4.2.3.1. Vehicles*

Within this section on specific traffic actors such vehicles, this is further split into section on vehicle lights, vehicle types (nominal and special), and state of these vehicle in its interaction with the ego.

##### *4.2.3.1.1. Vehicle Lights*

| **Example Category** | **Examples** | **Perception** | | **Situation Analysis** | | **Goal Management** | **Planning** |
|---|---|---|---|---|---|---|---|
| | | *-Position & State Estimation* | *-Detection & Tracking* | *-Context Awareness* | *-Prediction* | | |
| Intention Indicators | Brake/ Reverse/ Indicator | | *AV might have difficulty in classifying the state of lights on other vehicles.* | *If AV has difficulty in getting state of actor vehicle's lights, it could lead to difficulty to understand the situation and to define the adapted AV behaviour. E.g. did not realise emergency vehicle lights/ car ahead intention to reverse.* | *If AV has difficulty in getting state of actor vehicle's lights, it could lead to difficulty to predict actor vehicle's movement* | *If the AV is not able to understand the actor's intention, it will not be able to decide what is the action it should take in a safe way. i.e., understanding traffic flow versus right of way.* | *AV might have difficulty carrying out the planned manoeuvre due to wrong space estimation.* |
| Emergency vehicle lights | Flashing ambulance lights | | | | | | |

#### 4.2.3.1.2. *Types*

##### 4.2.3.1.2.1. ***Nominal***

| **Example Category** | **Examples** | **Perception** | | **Situation Analysis** | | **Goal Management** | **Planning** |
|---|---|---|---|---|---|---|---|
| | | *-Position & State Estimation* | *-Detection & Tracking* | *-Context Awareness* | *-Prediction* | | |
| Standard vehicles: | Heavy truck/ tow truck 3-wheel motorcycle/ truck/motorcycle/bicycle/bus/e-bike | | *AVs might have difficulty in classifying typical vehicles on SG roads due to the variety such as trucks/lorries.* | *Limitation of the perception capability can lead to performance limitation such incorrect understanding of traffic flow well. E.g. E-bike users do not undergo traffic rules training as stringently as motorcyclists and may often flout traffic rules. Cyclists sometimes switch roles between on-road 'vehicle' and footpath 'pedestrian' roles. E.g. big vehicles could be a source of occlusion.* | *Limitation of the perception capability can lead to performance limitation such incorrect prediction of traffic flow* | *If AV has difficulty perceiving information of actors, it could affect its balancing of goals between traffic flow and safety to other actors.* | *Limitation of the perception capability can lead to performance limitation such as incorrect estimation of space available.* |

***4.2.3.1.2.2. Special***

| Example Category | Examples | Perception | | Situation Analysis | | Goal Management | Planning |
|---|---|---|---|---|---|---|---|
| | | *-Position & State Estimation* | *-Detection & Tracking* | *-Context Awareness* | *-Prediction* | | |
| Emergency/ priority vehicles | Ambulance/ police/ fire truck | | *Difficulty in classification of emergency vehicles: police car, firetrucks, ambulance and associated lights.* | *AV may have difficulty in understanding the specific situation.*<br>*e.g., need to give way to emergency vehicles*<br>*e.g., overtaking road sweeper*<br>*e.g., the bicycle is not on the road surface, leading to possible*<br>*E.g. awareness to be keep a distance to funeral car.*<br>*E.g., Understand vehicle breakdown and being towed.* | *AV may have difficulty to predict the behaviour of the other road users that can understand the context of emergency vehicles or road sweepers.* | *AV may have difficulty in balancing goal management. To consider giving way to priority vehicle vs AV's own reachability and safety.* | *If AV has difficulty in context awareness of special vehicles, it may have difficulty in evaluate space available for intended manoeuvres.* |
| Road Sweeper | | | *Difficulty in classification of road sweeper as a unique type of road vehicle.* | | | *AV may have difficulty to manage reachability (following slow moving vehicle) versus safety when overtaking.* | |
| Modified (either add-on/ pictures) | Object on car; | | *Classification of atypical vehicles: Modification could affect classification of vehicle.*<br>*E.g., Modification such as a red-bull can on car as advertisement could lead to wrong classification of car to object*<br>*E.g. Could lead to wrong classification of bicycle instead of bus.* | | *AV may have difficulty to predict the movement of modified vehicles due to wrong or separate classification.*<br><br>*E.g., AVs prediction of towed vehicle could be wrong if using heading (facing direction).* | *If the AV is not able to understand the context, it might have difficulty managing goals of understanding traffic flow.* | |
| | Bicycle picture on bus | | | | | | |
| | *Funeral car* | | *AV may have difficulty in classification of different shape of vehicle. Could* | | | | |
| | | | *lead to generic classification as a vehicle.* | | | | |
| | *1.Lorry (open-deck) with objects sticking out*<br>*2. Vehicle being towed (in reverse)*<br>*3. Lorry (open-deck) with random objects within (classification)* | | *Difficulty for AV for classification due to shape mismatch from training database.* | | | | |
| Construction vehicles | Truck carrying loads/other vehicles | | *Difficulty in classification of vehicle. Extremities detected may not be classed together with the vehicle.* | | | | |

#### *4.2.3.1.3. Static State*

##### *4.2.3.1.3.1. Nominal*

| **Example Category** | **Examples** | **Perception** | | **Situation Analysis** | | **Goal Management** | **Planning** |
|---|---|---|---|---|---|---|---|
| | | *-Position & State Estimation* | *-Detection & Tracking* | *-Context Awareness* | *-Prediction* | | |
| Lanes allow for stopping of vehicles | 1. Lead vehicle stopped<br>2. Parked vehicle at side | | | *AV could have difficulty understand the nominal interactions with other vehicles and react* | *AV with difficulty in context awareness may result in* | *If AV has difficulty in context awareness, it could affect its balancing* | *If AV has difficulty in context awareness, it can lead to* |
| | of road.<br>3. Stopped vehicle for pick-up of drop off at roadside. | | | *accordingly (e.g. Awareness of parked vehicle (different from static vehicle in traffic).* | *incorrect prediction of actor movement.* | *of goals between traffic flow and safety to other actors.* | *performance limitation such as incorrect estimation of space available.* |

***4.2.3.1.3.2. <u>Behaviour against traffic rules</u>***

| **Example Category** | **Examples** | **Perception** | | **Situation Analysis** | | **Goal Management** | **Planning** |
|---|---|---|---|---|---|---|---|
| | | *-Position & State Estimation* | *-Detection & Tracking* | *-Context Awareness* | *-Prediction* | | |
| Lanes/ infrastructure restrict vehicles from stopping. | 1. Stopped vehicle obstructing the lane in an intersection<br>2. Vehicle parked at pedestrian crossing.<br>3. Vehicle stopped and picking up /drop off passengers at pedestrian crossing. | | | *AV could have difficulty understand the context of actor's intention. E.g., .AV could have difficulty in awareness of illegally parked vehicle E.g., understanding taxi carrying out an illegal pick-up /drop off and the action could take a while.* | *AV with difficulty in context awareness may result in incorrect prediction of actor movement.* | *If AV has difficulty context awareness of actors, it could affect its balancing of goals between traffic flow and safety to other actors.<br>E.g., if AV should wait behind or execute lane change* | *If AV has difficulty in context awareness, it can lead to performance limitation such as incorrect estimation of space available.<br>E.g., At intersection - space for safe lane change* |

#### 4.2.3.1.3.3. Visibility

| **Example Category** | **Examples** | **Perception** | | **Situation Analysis** | | **Goal Management** | **Planning** |
|---|---|---|---|---|---|---|---|
| | | *-Position & State Estimation* | *-Detection & Tracking* | *-Context Awareness* | *-Prediction* | | |
| Other users or infrastructures occluding the static vehicle | Parked vehicle occluded by a heavy truck in front of the EGO<br><br>Stopped vehicle in a discretionary right turn is occluded by a large vehicle at discretionary right turn waiting pocket | *AV is unable to have localisation information of occluded static vehicle.* | *AV might have difficulty detecting static vehicle. No information for tracking of static vehicle.* | *If AV can't perceive a vehicle in its environment it could lead to difficulty to understand the situation and to define the adapted AV behaviour* | *If AV can't perceive a vehicle in its environment it could lead to difficulty to predict the behaviour of this vehicle* | *If AV has difficulty in perceiving actors, it could affect its balancing of goals between right of way versus safety to other actors.* | *Visibility of static vehicle (no information) could affect AV's knowledge of available space to carry out the planned manoeuvre.* |

#### 4.2.3.1.4. *Dynamic State*

In this section, the dynamic vehicle is the actor element in the scenario, the interactions listed are between the ego vehicle and actor. The example category and interaction with ego are meant to be general, and the specific action of the actor or ego is listed within the example itself. The interactions are nominal, erratic, behaviour against traffic rules and visibility of dynamic vehicles.

##### 4.2.3.1.4.1. *Nominal*

<table>
<tr><th rowspan="2">Example Category</th><th rowspan="2">Interaction with Ego</th><th rowspan="2">Examples</th><th colspan="2">Perception</th><th colspan="2">Situation Analysis</th><th rowspan="2">Goal Management</th><th rowspan="2">Planning</th></tr>
<tr><th>-Position & State Estimation</th><th>-Detection & Tracking</th><th>-Context Awareness</th><th>-Prediction</th></tr>
<tr><td>Parallel</td><td>Lateral offset to ego</td><td>1. Dual Lane turning next to large vehicle</td><td></td><td></td><td rowspan="3">AV could have difficulty understand the nominal interactions with other vehicles and react accordingly.<br>E.g., Parallel: Awareness that large vehicle needs different turning radius. Lateral clearance should not be compromised.<br>E.g., Lane Change: Understanding which lane gives AV the right of way and space for manoeuvre while anticipating traffic interaction.<br>E.g., Oncoming: Understanding actor intention and AV’s right of way.</td><td rowspan="3">AV with difficulty in context awareness may result in incorrect prediction of actor movement.</td><td rowspan="3">AV could have difficulty balancing of goals between maintaining right of way and traffic flow and safety to other actors.<br>E.g., Managing safe lateral clearance should not be compromised, slow down and stay behind vehicle or speed up.</td><td rowspan="3">If AV has difficulty in context awareness. It can lead to performance limitation such as incorrect estimation of space available.</td></tr>
<tr><td>Lane Change</td><td>Lane Change</td><td>1. Actor lane change into AV lane; (ego being overtaken)<br>2. AV changing lane ahead of actor in adjacent lane</td><td></td><td></td></tr>
<tr><td>Collision point interactions/ Crossings</td><td>From side executing Turns (left, right, U-turn)</td><td>1. Ego going straight, actor turning into ego path - left turn slip road/ coming out from minor road/discretionary right turn/U-turn<br>2. Ego executing turns - judging safe time/distance to actor - (lane</td><td></td><td></td></tr>
</table>

| | | | | | | | | |
|---|---|---|---|---|---|---|---|---|
| | | change behaviour expected)<br>3. Ego & actor turning into same road. | | | | | | |
| | Oncoming | 1. Ego going straight, actor overtaking vehicle its lane and on ego's lane (safe distance away)<br>2. Ego overtaking and going into adjacent lane with oncoming actor (safe distance away).<br>3. Vehicle ahead reversing for parking. | | | | | | |
| Following | Actor behind ego | nominal traffic flow[5] | | | | | *AV could have difficulty managing traffic flow behind it when carrying out other goals from interaction with other actors.* | *AV could have difficulty in deciding manoeuvre to manage the space available to actor behind.* |

[5] This management of traffic flow behind an AV is a general consideration that should always be taken into account in goal management. It is highlighted here for specific scenarios creation that wants to represent traffic flow behind an AV explicitly.

### 4.2.3.1.4.2. ***Erratic***

| **Example Category** | **Interaction with Ego** | **Examples** | **Perception** | | **Situation Analysis** | | **Goal Management** | **Planning** |
|---|---|---|---|---|---|---|---|---|
| | | | *-Position & State Estimation* | *-Detection & Tracking* | *-Context Awareness* | *-Prediction* | | |
| Parallel | Actor not keeping in lane | 1. Actor weaving in adjacent lane. (could be losing control)<br>2. Dual lane turning, actor in adjacent lane drifting into AV's lane.<br>3. Actor lane splitting (motorcycle/ e-bike) | | | *AV could have difficulty understand the context of actor's intention.*<br>*E.g., Loss of control could be mis-interpreted as sudden lane change*<br>*E.g. motorcycle lane splitting might suddenly cut into lane.*<br>*E.g., Cyclists sometimes change their path - footpath/ road. Switch between pedestrian and vehicle.*<br>*E.g., Bicycle usually assumes right of way, it might cross at a high speed and may not slow down approaching pedestrian crossing.* | *Sudden and unusual behaviours of the Actors (here moving vehicle) are difficult to predict.* | *Each interaction with other erratic vehicles affects AV's goal of driving in lane and sudden need to maintain safe distance and traffic flow.* | *AV could have difficulty in estimating available space for manoeuvre to react to actor's erratic movement.* |
| Lane Change | Cutting into AV's lane | 1. Actor in adjacent lane, abrupt lane change.<br>2. Bicycle - from pedestrian crossing become on-road 'actor' suddenly. | | | | | | |
| | Cutting out of AV's lane | 1. Actor sudden lane change (cutting out) | | | | | | |
| Following | Ego is following the actor | 1. Actor sudden braking ahead of ego<br>2. Actor weaving ahead of ego within lane. (Could be losing control) | | | | | | |
| | Actor following ego | 1. Actor behind is tailgating ego (very close to ego) | | | | | | |
| Collision point interactions/ Crossings | From side, executing turns | 1. Ego going straight, actor turning into ego path but stopping strangely - left turn slip road/ coming out from minor road/U-turn<br>2. Bicycle approaching very fast to a pedestrian crossing. | | | | | | |
| | Oncoming | Actor oncoming, overtaking parked cars, very close in approach. | | | | | | |

### *4.2.3.1.4.3. Behaviour against traffic rules*

| **Example Category** | **Interaction with Ego** | **Examples** | **Perception** | | **Situation Analysis** | | **Goal Management** | **Planning** |
|---|---|---|---|---|---|---|---|---|
| | | | *-Position & State Estimation* | *-Detection & Tracking* | *-Context Awareness* | *-Prediction* | | |
| Lane Change | | Actor changing lane across double white line into ego's lane. | | | *AV is not able to understand the traffic context and react accordingly to the situation.* | *Sudden and unusual behaviour of the actors are difficult to predict.* | *Each interaction with other illegal vehicles affects AV's goal of right of way versus safety to other actors.* | *Affects AV's available space since this is violated with actor's illegal manoeuvres.* |
| Collision point interactions/ Crossings | Oncoming | Actor heading towards ego in ego's lane while:<br>1. overtaking vehicle crossing double white line<br>2 actor is travelling on the wrong lane. | | | | | | |
| | From side | Actor beating a red light (right/ left/ straight from left or right) | | | | | | |

#### 4.2.3.1.4.4. *Visibility*

| Example Category | Examples | Perception | | Situation Analysis | | Goal Management | Planning |
|---|---|---|---|---|---|---|---|
| | | *-Position & State Estimation* | *-Detection & Tracking* | *-Context Awareness* | *-Prediction* | | |
| Other users or infrastructures occluding dynamic vehicle | Plastic tarp flies off prevents ego view of traffic ahead<br>Barrier blocking low car/motorcycle moving<br>Piling machine occluding view of traffic flow | *AV is unable to have localisation information of occluded dynamic vehicle.* | *Unable to detect dynamic vehicle. No information for tracking of dynamic vehicles* | *If AV can't perceive a vehicle in its environment, it could lead to difficulty to understand the situation and to define the adapted AV behaviour.* | *If AV can't see a vehicle in its environment, it could lead to difficulty to predict the motion of this vehicle* | *Visibility of dynamic vehicle affects balancing of goal management of traffic flow versus safety. If the AV is not able to see the dynamic vehicle it will not be able to decide what is the action it should take in a safe way.* | *Visibility of dynamic vehicle) affects AV's knowledge of available space to carry out the planned manoeuvre.* |

### 4.2.3.2. Pedestrian

#### 4.2.3.2.1. Types

| Example Category | Examples | Perception | | Situation Analysis | | Goal Management | Planning |
|---|---|---|---|---|---|---|---|
| | | *-Position & State Estimation* | *-Detection & Tracking* | *-Context Awareness* | *-Prediction* | | |
| With accessories | Stroller/ Shopping trolley/ Umbrella/ Elderly with walking aids/ Pedestrian with phone | | *AV could have difficulty in classification and may not recognise accessories in addition to pedestrian.* | *AV could have difficulty understand the context of actor's movement and adapt its behaviour, especially if classification is wrong/ misdetection. E.g., If stroller/walking aids are not identified, AV does not have awareness of danger of child in stroller or elderly's difficulty in movement. E.g. child is erratic, elderly is slow.* | *AV with difficulty in context awareness may result in incorrect prediction of actor movement. E.g., Child's/elderly classification could affect prediction trajectory characteristic. E.g., misdetection of pedestrian could lead to difficulty to predict the motion of the pedestrian.* | *If AV has difficulty perceiving information of actors, it could affect AV's balancing of goals between traffic flow and safety to other actors. E.g., more time/anticipation for VRU movement.* | *Misdetection / classification of a pedestrian affects knowledge of available space to carry out the planned manoeuvre for an AV.* |
| Nominal Standard | Child; Elderly (behaviour- cautious) | | *AV could have difficulty in classification for child/elderly as a pedestrian due to its size or pose.* | | | | |
| Nominal- detectability | Pedestrian far away; close to infrastructure | | *AV could have difficulty in classification. E.g. when pedestrians are close to other infrastructure and might not be detected separately, E.g., Too far away to be classified as a pedestrian* | | | | |

#### 4.2.3.2.2. *Static State*

##### 4.2.3.2.2.1. *Nominal*

| Example Category | Examples | Perception | | Situation Analysis | | Goal Management | Planning |
|---|---|---|---|---|---|---|---|
| | | *-Position & State Estimation* | *-Detection & Tracking* | *-Context Awareness* | *-Prediction* | | |
| Away from 'pedestrian crossing' infrastructure | Close to kerb. | | | *AV could have difficulty to understand the intention of pedestrian's and to react accordingly to the situation. E.g., Such as pedestrians standing further back do intend to cross but waiting further back.* | *AV could have difficulty to predict potential pedestrian movement, especially if context awareness is not achieved. E.g., Right of way of pedestrian* | *AV could have difficulty balancing of goals between traffic flow and safety to other actors.* | *AV could have difficulty in estimating space available for manoeuvre due to prediction of pedestrian movement* |
| Waiting at pedestrian crossing infrastructure. | Standing under shade waiting to cross | | | | | | |

##### 4.2.3.2.2.2. *Behaviour against traffic rules*

| Example Category | Examples | Perception | | Situation Analysis | | Goal Management | Planning |
|---|---|---|---|---|---|---|---|
| | | *-Position & State Estimation* | *-Detection & Tracking* | *-Context Awareness* | *-Prediction* | | |
| Standing on road | 1.Pedestrian on road in lane<br>2. Pedestrian crossed late and stopped at traffic island. | | | *AV could have difficulty to understand the traffic context when unexpected behavior against traffic rules happen and react accordingly to the situation.* | *AV could have difficulty in predicting future pedestrian movements. E.g., Is it planning to move in the next moment?* | *AV could have difficulty balancing of goals between right of way versus safety to other actors.<br>E.g., Allow pedestrian to finish crossing (BTD 146) even though AV has right of way.* | *AV could have difficulty in estimating space available for manoeuvre due erratic pedestrian position.* |

##### 4.2.3.2.2.3. Visibility

| Example Category | Examples | Perception | | Situation Analysis | | Goal Management | Planning |
|---|---|---|---|---|---|---|---|
| | | *-Position & State Estimation* | *-Detection & Tracking* | *-Context Awareness* | *-Prediction* | | |
| Other users or infrastructures occluding pedestrian | Pedestrians occluded by bus stop infrastructure (pillar or advertisement panel) /Static pedestrian occluded by a large truck in front of the Ego | *AV is unable to have localisation information of occluded road users.* | *Unable to detect road users. No information for tracking other road users* | *If AV can't perceive a pedestrian in its environment it could lead to difficulty to understand the situation and to define the adapted AV behaviour towards pedestrians.* | *Visibility of pedestrian affects predictability. If the AV is not able to see pedestrians, it will not be able to predict the future movement and to react accordingly to the situation* | *Visibility of pedestrian affects balancing of goal. If the AV is not able to see the pedestrians, it will not be able to decide what is the action it should take in a safe way.* | *Visibility of static pedestrian could affect AV's knowledge of available space to carry out the planned manoeuvre.* |

#### 4.2.3.2.3. Dynamic State

##### 4.2.3.2.3.1. Nominal

| Example Category | Examples | Perception | | Situation Analysis | | Goal Management | Planning |
|---|---|---|---|---|---|---|---|
| | | *-Position & State Estimation* | *-Detection & Tracking* | *-Context Awareness* | *-Prediction* | | |
| Traffic lights; /Pedestrian crossing /Non-pedestrian crossing | 1. Pedestrian crossing - following lights at traffic light<br>2. Pedestrian crossing at zebra crossing<br>3. Pedestrian crossing road (non-jaywalking areas) | | | *AV could have difficulty to understand the intention of pedestrians in nominal situations* | *AV could have difficulty in predicting future pedestrian movements if it has difficulty understanding context.* | *AV could have difficulty balancing of goals between right of way versus safety to other actors.*<br>*E.g., AV should provide to pedestrian priority and right of way if applicable.* | *AV could have difficulty in estimating space available for manoeuvre while anticipating traffic interaction.*<br>*E.g., 3. AV should be able to detect pedestrians are crossing (safe distance ahead) and react if slower speed is required.* |

#### 4.2.3.2.3.2. ***Behaviour against traffic rules***

| Example Category | Examples | Perception | | Situation Analysis | | Goal Management | Planning |
|---|---|---|---|---|---|---|---|
| | | *-Position & State Estimation* | *-Detection & Tracking* | *-Context Awareness* | *-Prediction* | | |
| Pedestrian crossing | Pedestrian crossing road late/early relative to traffic light; | | | *Difficulty for AV to be aware of pedestrians with unexpected movements* | *AV may have difficulty in predicting pedestrian movement* | *AV could have difficulty balancing of goals between traffic flow/right of way and safety to other actors.* | *AV could have difficulty estimating available space since this is violated with unexpected pedestrian behaviour.* |
| Non-pedestrian crossing | Pedestrian Jaywalking/ | | | | | | |

#### 4.2.3.2.3.3. ***Erratic***

| Example Category | Examples | Perception | | Situation Analysis | | Goal Management | Planning |
|---|---|---|---|---|---|---|---|
| | | *-Position & State Estimation* | *-Detection & Tracking* | *-Context Awareness* | *-Prediction* | | |
| Not minding traffic | Pedestrian using phone and suddenly crossing. /Children running/ approaching crossing - sudden change in speeds/direction /Cyclists speeding across zebra crossing assuming right of way | | | *AV may have difficulty in understanding of context awareness of various pedestrian intention and behaviour.*<br>*E.g., Understanding that pedestrian would need time to recover after falling down.*<br>*E.g., AV may not be aware of pedestrian's distraction with phone.* | *Sudden and unusual behaviours of the Actors are difficult to predict.* | *AV could have difficulty balancing of goals between traffic flow and safety to other actors* | *AV could have difficulty estimating available space since this is violated with actor's erratic movement.* |
| Change of mind | Pedestrians hesitant at zebra crossing; | | | | | | |
| Others | Pedestrian falling when crossing /Pedestrian gesturing (signal danger of child running behind parked car) | | *AV could have difficulty in classification of pedestrian in a different orientation.* | *E.g., pedestrian gesturing could imply danger of another actor.* | | | |

#### *4.2.3.2.3.4.* ***Visibility***

| **Example Category** | **Examples** | **Perception** | | **Situation Analysis** | | **Goal Management** | **Planning** |
|---|---|---|---|---|---|---|---|
| | | *-Position & State Estimation* | *-Detection & Tracking* | *-Context Awareness* | *-Prediction* | | |
| Other users or infrastructures occluding dynamic pedestrian | Pedestrian crossing road occluded by a large truck in front of AV. | *AV is unable to have localisation information of occluded road users.* | *Unable to detect road users. No information for tracking other road users* | *If AV can't perceive a pedestrian in its environment it could lead to difficulty to understand the situation and to define the adapted AV behaviour towards pedestrian.* | *Visibility of pedestrian affects predictability. If the AV is not able to perceive pedestrians, it will not be able to predict the future movement and to react accordingly to the situation* | *Visibility of pedestrian affects AV's balancing of goals. If the AV is not able to see the pedestrians, it will not be able to decide what is the action it should take in a safe way.* | *Visibility of dynamic pedestrian could affect AV's knowledge of available space to carry out the planned manoeuvre.* |

## 4.2.3.3. *Object (near/on road surface)*

### 4.2.3.3.1. *Static State*

#### 4.2.3.3.1.1. ***Permanent***

| Example Category | Examples | Perception | | Situation Analysis | | Goal Management | Planning |
|---|---|---|---|---|---|---|---|
| | | *-Position & State Estimation* | *-Detection & Tracking* | *-Context Awareness* | *-Prediction* | | |
| Roadside vegetation | Vegetation on roadside that needs trimming | | *AV may have difficulty classifying/ identifying branches of shrubs/vegetations extension onto driveable area.* | *AV may not be aware if object can be passable.* | *AV may difficulty in predicting other road user's behaviour due to their understanding of the object.* | *Each interaction with object affects AV's goal of driving in lane and need to maintain safe distance and traffic flow. (e.g., unnecessary lane change)* | *This affects AV's space for manoeuvre due to wrong object classification and drivable area.* |
| Advertisement | 1. Petrol station advertisement (air-filled moving balloon)<br>2. Restaurant signs | | *AV might have difficulty in classification.*<br>*1. Misclassification as human*<br>*2. Misclassification of restaurant signs as traffic signs.* | *AV may have difficulty in understanding object and wrongly identifies an expected behaviour to the misclassified object.*<br>*1. Context of advertisement and not a pedestrian potential movement*<br>*2. Information of restaurant interpreted as roadworks and creates a warning in AV system.* | | | |

### 4.2.3.3.1.2. ***Temporary***

| **Example Category** | **Examples** | **Perception** | | **Situation Analysis** | | **Goal Management** | **Planning** |
|---|---|---|---|---|---|---|---|
| | | *-Position & State Estimation* | *-Detection & Tracking* | *-Context Awareness* | *-Prediction* | | |
| Others | Box, papers | | *AV may have difficulty classifying/ identifying unusual object on the road* | *AV may have difficulty to understand the situation (if the object is passable) and to define the adapted AV behaviour when an object is on the road.* | *AV may have difficulty in predicting other road user's behaviour due to their understanding of the object.* | *Each interaction with object affects AV's goal of driving in lane and need to maintain safe distance and traffic flow. (Unnecessary lane change)* | *This affects AV's space for manoeuvre due to wrong object classification and drivable area.* |
| Natural object | Dead animals on road; /Stones/leaves on road surface | | | | | | |

### 4.2.3.3.1.3. ***Visibility***

| **Example Category** | **Examples** | **Perception** | | **Situation Analysis** | | **Goal Management** | **Planning** |
|---|---|---|---|---|---|---|---|
| | | *-Position & State Estimation* | *-Detection & Tracking* | *-Context Awareness* | *-Prediction* | | |
| Other users or infrastructures occluding static object | Vehicle ahead blocking view of dead animal on road. | *AV is unable to have localisation information of occluded object.* | *Unable to detect object. Does not have information for tracking.* | *If AV can't see the object, it could lead to difficulty to understand the situation and to define the adapted AV behaviour E.g., Difficulty in understanding behaviour of vehicle ahead to avoid objects that are occluded. AV may not take it as a sign of danger.* | *If AV can't perceive an object in its environment, it could lead to difficulty to predict the potential motion of this static object.* | *Visibility of static object affects the balancing of goals. AV does not realise it could have the situation of managing static object in AV's path, to manage between continuing driving in lane (passable) or to change lane.* | *Visibility of static object could affect AV's knowledge of available space to carry out the planned manoeuvre.* |

#### *4.2.3.3.2. Dynamic State*

##### *4.2.3.3.2.1. Temporary*

| Example Category | Examples | Perception | | Situation Analysis | | Goal Management | Planning |
|---|---|---|---|---|---|---|---|
| | | *-Position & State Estimation* | *-Detection & Tracking* | *-Context Awareness* | *-Prediction* | | |
| Objects falling off vehicle | Runaway tires; boxes /Panels flying off *(non-secured loads)* | | *AV may have difficulty classifying/ identifying unusual object on the road* | *AV may have difficulty to understand the situation and to define the adapted AV behaviour when an object is moving on the road, and if the object is passable.* | *AV might have difficulty on predicting movements of the object.* | *Each interaction with an object affects AV's goal of driving in lane and need to maintain safe distance versus traffic flow.* | *Difficulty to understand the situation and to define the adapted AV maneuver when an object is moving on the road* |
| Object from pedestrian | Bouncing balls | | | | | | |
| Animals crossing road | Chicken, bird /Wild boar | | | | | | |

##### *4.2.3.3.2.2. Visibility*

| Example Category | Examples | Perception | | Situation Analysis | | Goal Management | Planning |
|---|---|---|---|---|---|---|---|
| | | *-Position & State Estimation* | *-Detection & Tracking* | *-Context Awareness* | *-Prediction* | | |
| Other users or infrastructures occluding dynamic object | 1. Vehicle ahead/ adjacent lane occluding view of runaway tires.<br>2. Vehicle occluding view of animal crossing road | *AV is unable to have localisation information of occluded dynamic object.* | *AV is unable to detect road object. No information for tracking.* | *If AV can't see the object, it could lead to difficulty to understand the situation and to define the adapted AV behaviour<br>E.g., Difficulty in understanding behaviour of vehicle ahead to avoid objects that are occluded. Does not take it as a sign of danger.* | *If AV can't see the object, it could lead to difficulty on predicting movements of the object.* | *Visibility of dynamic object affects goal management. Does not realise it could have the situation of managing object in AV's path, to manage between continuing driving in lane (passable) or to change lane.* | *Visibility of dynamic object affects AV's estimation of space available for manoeuvre.* |

### 4.2.4. Environmental Conditions

This refers to the environmental conditions that could happen *temporarily* during a drive.

#### *4.2.4.1. Visibility*

| **Example Category** | **Examples** | **Perception** | | **Situation Analysis** | | **Goal Management** | **Planning** |
|---|---|---|---|---|---|---|---|
| | | *-Position & State Estimation* | *-Detection & Tracking* | *-Context Awareness* | *-Prediction* | | |
| Weather conditions | Rain, storm, fog | *Limitation of the perception capability can lead to performance limitation such incorrect measurements E.g., A glare could interfere with the laser detection, leading to incorrect measurements of distance and location of the objects on the vehicle's surroundings.* | *Limitation of the perception capability can lead to performance limitation such incorrect classification, incorrect tracking, misdetection. The accuracy and range of detection of the sensors are strongly affected. The system might not be able to detect objects with the same accuracy as it would be in more favorable conditions. E.g., in low light conditions, objects might not reflect enough light, leading to reduced detection accuracy for a Lidar. Cameras might not be able to capture enough light to create a detailed accurate image.* | *Limitation of the perception capability can lead to an AV's incorrect understanding of traffic flow as they might not be able to detect accurately obstacles, pedestrian, or other vehicles in their path.* | *Limitation of the perception capability can lead to incorrect prediction of traffic flow* | *This affects goal management of driving safely in the ODD. The AV may have difficulty in evaluation of the confidence of visibility. E.g., balance the aware of sensor limitation in rain versus reachability.* | *Limitation of the perception capability can lead to performance limitation such as incorrect estimation of space available.* |
| Suspended particles | Smoke, dust, haze. | | | | | | |
| Illumination - Glare Lighting | Glare (low angled sun, oncoming vehicles headlights, public lighting) | | | | | | |
| Illumination – Low levels | Dusk or dawn /Night with dim public light on /Night with public lights off /Cloudiness level /Shadow | | | | | | |

#### 4.2.4.2. Others

| Example Category | Examples | Perception | | Situation Analysis | | Goal Management | Planning |
|---|---|---|---|---|---|---|---|
| | | *-Position & State Estimation* | *-Detection & Tracking* | *-Context Awareness* | *-Prediction* | | |
| Weather induced road conditions | wet pavement, flooded, pavement, water puddle | | *AV might have difficulty detecting lanes due to reflective nature of road conditions.* | *AV may not have awareness to understand the ego's capability for these road conditions and weather elements to adapt its behaviour accordingly.<br>E.g., If the conditions of the road are not detected by the AV, AV will not have the awareness needed to adjust its behaviour (reduce speed, modify trajectory)<br>E.g., Strong winds could also affect the stability of the vehicle causing the deviation of its programmed trajectory. Awareness this condition and the Ego's capability are needed to counteract these effects ensuring safe operations in windy conditions.* | *Limitation of the perception capability can lead to incorrect prediction of traffic flow* | *Could be difficult to achieve AV's reachability (changing lane /rerouting) versus safety. Vehicle's speed/braking strategy for tackling road conditions.* | *Road condition could affect space where manoeuvres are to be carried out in.* |
| Weather conditions (not affecting the visibility) | Wind | | *AV might have difficulty detecting windy conditions.* | | | | |

# 5. Analysis of identified accidents or incidents

The second part of the approach is to highlight the usage of the guidelines created above to evaluate accidents or incidents that occur on Singapore roads from Resembler's webtool [13]. The accidents or incidents are chosen to cover a variety of difficulties for AVs such as occlusion of actors, roadworks, unpredictable human behaviour such as erratic VRU movements at road crossings or interaction with erratic human-driven vehicles. These accidents and incidents are analysed to demonstrate the use of guidelines in understanding how on-road traffic interactions could be broken down to the corresponding complexity components for AV trials on road.

**Overview of analysis method:**

The method used for analysis of chosen accidents and incidents (events) on Singapore roads to understand them as complex scenarios for AVs is as follows:

1) Identify traffic agent hazard elements from guidelines observed in the accident or incident, then consolidate the list of corresponding complexity components associated.
2) Overall analysis of each element of traffic agent hazard within the context of the accident or incident.
3) Provide the final analysis of the accident overall, where the accident or incident could be an occurrence of traffic situation that an AV might encounter on Singapore roads.

In the selection of accidents and incidents from database from Resembler webtool, five events were chosen for analysis, covering the variety of elements of traffic agent hazards to highlight the occurrence of such situations in real traffic that are complex for an AV. The following list are the names of the accident or incident selected for analysis.

1. Pedestrian jaywalking with stroller.
2. Traffic junction with VRU not following rules.
3. Work zone and sudden lane change.
4. Reachability with traffic flow.
5. Sudden cut out and sudden braking.

## 5.1. Example 1: Pedestrian jaywalking with stroller

This recorded incident occurred on an urban two-lane road where an AV could interact with pedestrians crossing the road. In this incident, traffic on the right lane comes to a standstill due to a traffic red light ahead whereas the left lane remains empty. A lady is crossing the stroller from the opposite side of the road on the right and enters onto the empty left lane in the path of the vehicle with the recording dash camera without hesitation. The driver does not see this lady with the stroller due to occlusion of them behind the vehicles on the right lane but is able to brake and stop the vehicle in time when the stroller wheels were first visible. The lady continues to emerge from the occlusion, only stopping when the stroller and herself are within the left lane, however as the vehicle is fully stopped, the lady continues to cross the road and moves towards the left edge of the vehicle. From this accident, we will evaluate the complexity for an AV in the following analysis considering the AV as the camera vehicle.

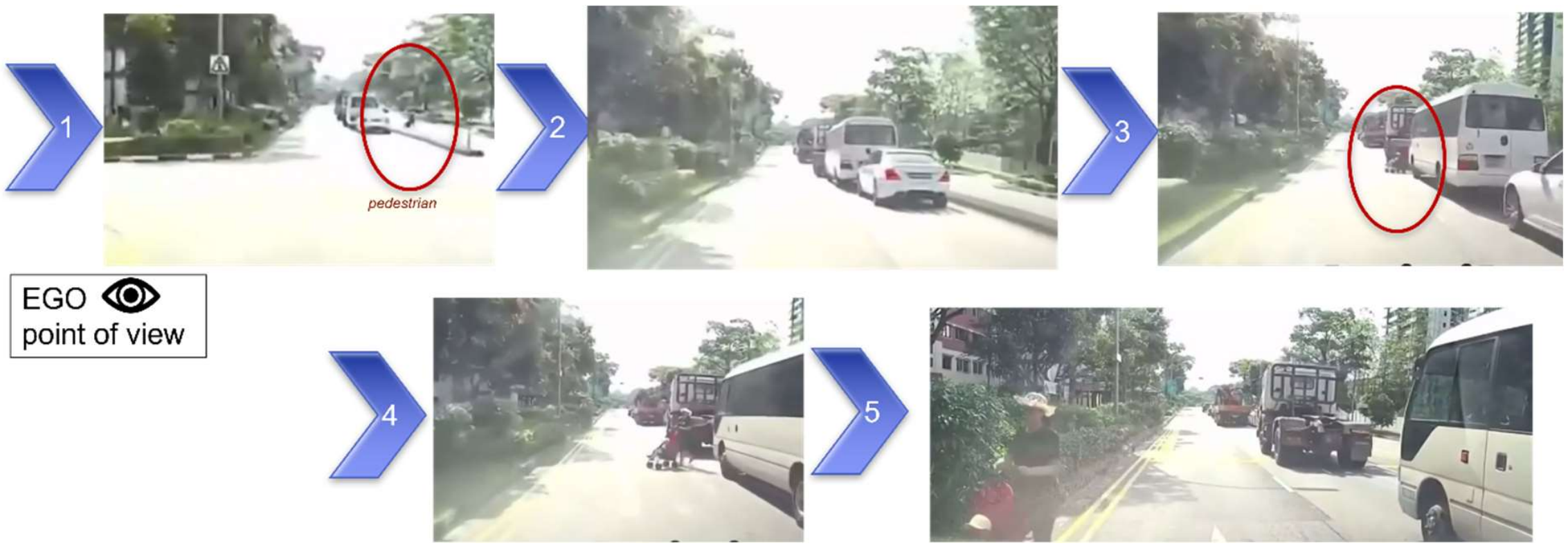

*Figure 3: Images from video recording of an incident where pedestrian with stroller crossed a road [13].*

### 5.1.1. Identify traffic agent hazard elements from guidelines and its associated complexity.

The following elements of traffic agent hazards were identified from this incident recorded. The assumption for a complex scenario from this recorded incident is that an AV would be in the point of view of the vehicle with the dashcam recording.

- **Traffic Actor: Vehicle**
  - Type - Nominal
  - State -Static
    - Nominal
- **Traffic Actor: Pedestrian**
  - Type- With accessories
  - Type- Nominal - Detectability
  - State - Dynamic
    - Nominal (opposite side of the road)
    - Visibility
    - Behaviour against traffic rules

The complexity of these elements is taken from the guidelines and compiled in the table below:

Table 1: *Elements of traffic agent hazards in this incident and the associated complexity from the guidelines above.*

| **Elements of Traffic Agent Hazards** | | ***Situation Awareness*** | | | | **Decision Making** | |
|---|---|---|---|---|---|---|---|
| Traffic Agent Hazards | Details of traffic agent hazards | **Perception** | | **Situation Analysis** | | **Goal Management** | **Planning** |
| | | *-Position & State Estimation* | *-Detection & Tracking* | *-Context Awareness* | *-Prediction* | | |
| Actor: Pedestrian | Dynamic State; Nominal, non-pedestrian crossing | | | *AV could have difficulty to understand the intention of pedestrians in nominal situations* | *AV could have difficulty in predicting future pedestrian movements if it has difficulty understandin g context.* | *AV could have difficulty balancing of goals between right of way versus safety to other actors. E.g., AV should provide to pedestrian priority and right* | *AV could have difficulty in estimating space available for manoeuvre while anticipating traffic interaction.* |
| | | | | | | *of way if applicable.* | *E.g., 3. AV should be able to detect pedestrians are crossing (safe distance ahead) and react if slower speed is required.* |
| | Type – Nominal Detectability | | *AV could have difficulty in classification E.g., Too far away to be classified as a pedestrian* | *AV could have difficulty understand the context of actor's movement and adapt its behaviour, especially if classification is wrong/ misdetection.* | *AV with difficulty in context awareness may result in incorrect prediction of actor movement.* | *If AV has difficulty perceiving information of actors, it could affect AV's balancing of goals between traffic flow and safety to other actors.* | *Misdetection / classification of a pedestrian affects knowledge of available space to carry out the planned manoeuvre for an AV.* |
| | Type – With accessories | | *AV could have difficulty in classification and may not recognise accessories in addition to pedestrian.* | *AV could have difficulty understand the context of actor's movement and adapt its behaviour, especially if classification is wrong/ misdetection. E.g., If stroller/walking aids are not identified, AV does not have awareness of danger of child in stroller or elderly's difficulty in movement.* | *AV with difficulty in context awareness may result in incorrect prediction of actor movement. E.g., misdetection of pedestrian could lead to difficulty to predict the motion of the pedestrian.* | *If AV has difficulty perceiving information of actors, it could affect AV's balancing of goals between traffic flow and safety to other actors.*<br><br>*E.g., more time/anticipation for VRU movement.* | *Misdetection / classification of a pedestrian affects knowledge of available space to carry out the planned manoeuvre for an AV.* |
| | Dynamic State-Visibility | *AV is unable to have localisation information of occluded road users.* | *Unable to detect road users. No information for tracking other road users* | *If AV can't perceive a pedestrian in its environment it could lead to difficulty to understand the situation and to* | *Visibility of pedestrian affects predictability. If the AV is not able to see pedestrians,* | *Visibility of pedestrian (not in AV path) affects AV's goal management. If the AV is not able to see the pedestrians, it* | *Visibility of dynamic pedestrian could affect AV's knowledg* |
| | | | | *define the adapted AV behaviour towards pedestrian.* | *it will not be able to predict the future movement and to react accordingly to the situation* | *will not be able to decide what is the action it should take in a safe way.* | *e of available space to carry out the planned manoeuvre.* |
| | Dynamic State- Behaviour against traffic rules non-pedestrian crossing | | | *Difficulty for AV to be aware of pedestrians with unexpected movements* | *Difficulty in predicting pedestrian movement* | *Safety to VRU versus traffic flow. Each interaction with other illegal pedestrian crossing affects AV's goal of driving in lane and sudden need to find available space to maintain safety distance (avoid collision). Traffic flow may be of lower priority* | *Affects AV's available space since this is violated with illegal crossing.* |
| Actor: Vehicle | Type- Nominal | | *AVs might have difficulty in classifying typical vehicles on SG roads due to the variety such as trucks/lorries.* | *Limitation of the perception capability can lead to performance limitation such incorrect understanding of traffic flow well. E.g., big vehicles could be a source of occlusion.* | *Limitation of the perception capability can lead to performance limitation such incorrect prediction of traffic flow* | *If AV has difficulty perceiving information of actors, it could affect its balancing of goals between traffic flow and safety to other actors.* | *Limitation of the perception capability can lead to performance limitation such as incorrect estimation of space available.* |
| | Static State – Nominal Lanes allow for stopping vehicle | | | *AV could have difficulty understand the nominal interactions with other vehicles and react accordingly (e.g. Awareness of parked vehicle (different from static vehicle in traffic).* | *AV with difficulty in context awareness may result in incorrect prediction of actor movement.* | *If AV has difficulty in context awareness, it could affect its balancing of goals between traffic flow and safety to other actors.* | *If AV has difficulty in context awareness, it can lead to performance limitation such as incorrect estimation of space available.* |

According to the Road Traffic Act on pedestrian crossings "13.—(1) Every pedestrian, cyclist, mobility vehicle user and Personal Mobility Device (PMD) rider, when crossing a road shall do so by the most direct route to the opposite side, and when crossing at any place other than a pedestrian crossing shall yield the right of way to all vehicles" [14]. In this event, the pedestrian has not checked for vehicle moving in the lane of the recording car and not yielded right of way by entering the lane, hence it is under the classification of 'behaviour against traffic rules at a non-pedestrian crossing'. When the pedestrian was initially crossing the road on the opposite road direction, there were no vehicles and hence can be initially classified as 'nominal' for a dynamic pedestrian, because the pedestrian was not hindering traffic flow in terms of the road traffic act above.

### 5.1.2. Overall analysis of each element in the context of the scenario

| Element | Element Category | Complexity Category | Complexity Reason |
|---|---|---|---|
| Pedestrian | Dynamic – Nominal *(nominal crossing outside pedestrian traffic light) started crossing on opposite side of road* | **Tracking; Context awareness** | Complexity of awareness of pedestrian crossing on opposite side of road, may continue to cross onto AV path, difficulty in tracking by an AV.<br><br>If the Ego is able to perceive the stroller and pedestrian at this moment (before entering in the occlusion of the stopped vehicles), this will give him the awareness about the possibility that the stroller could appear between the vehicles later. The initial recognition and tracking of this element may have provided a pre recognition of the future danger. |
| Pedestrian | With accessories | **Detection and Tracking** | Complexity of detection of stroller: AV may not recognize/classify stroller and related danger. |
| *Pedestrian* | Dynamic -Visibility *+With accessories (stroller halfway out behind vehicle)* | **Detection; Context awareness** | Complexity of detection: AV does not recognize half a stroller and the imminent danger *(stroller--> child in stroller --> pushed by pedestrian --> pedestrian is occluded)* |
| | | **Prediction:** | Complexity of prediction of stroller and pedestrian since it was occluded by car |
| | | **Goal Management Planning** | Occlusion of dynamic pedestrian affects AV's goal management. If the AV is not able to perceive the dynamic pedestrian, it will not be able to decide what is the action it should take in a safe way considering traffic flow. This could lead to chain incident with the vehicles behind the ego. |
| Pedestrian | Dynamic - Behaviour against traffic rules | **Prediction** | Difficulty to predict unusual behaviour of pedestrian |
| | | **Goal Management** | Complexity in managing goals between keeping safety to VRU versus maintaining traffic flow. Each interaction with other illegal pedestrian crossing affects AV's goal of driving in lane and sudden need to find available space to maintain safety distance (avoid collision). Traffic flow may be of lower priority |
| Vehicle- static | | **Context awareness** | Complexity of awareness of possible hidden actors behind. |

### 5.1.3. Final analysis of scenario complexity

In this incident where no collision occurred, this could be a scenario where an AV could encounter on public roads. The pedestrian with a stroller could have been detected in the initial crossing on the opposite side of the road, this initial detection of the pedestrian may have provided a pre-recognition of the future danger increasing the situation awareness and facilitating the prediction of possible future crossing of the pedestrian.

The first difficulty for the AV is to detect and classify from far the pedestrian being able to predict that the pedestrian could continue onto its lane. Secondly some AVs might have difficulty in classification of a stroller and the associated dangers such that a child is in the stroller that is the situation awareness functionality. The stroller with pedestrian may not be sufficiently presented in the training data for the perception system.

A third difficulty will be due to the occlusion of the pedestrian and stroller behind the static vehicle, some AVs may not have the capability to track this user continuously during the time it is occluded by the static vehicle on the right lane and predict its movement, even though initial detection was achieved when the pedestrian and stroller were initially crossing on the opposite side of the road. This overall leads to difficulty in prediction of the pedestrian with the stroller possibly coming out between the static vehicles (after occlusion).

If an AV was able to have the contextual awareness of the pedestrian with stroller going to cross from behind the parked vehicle (either through its capability of tracking or understanding by other methods the possibility of an occluded pedestrian could be behind the stopped vehicles), it could still face a difficulty in management of goals of maintaining traffic flow behind itself and safety to the pedestrian when it infringes on an AV’s drivable lane area. This would mean the AV needs to have the awareness of these goals and manage its planning movement such as speed so that it would not have to brake heavily if the pedestrian crosses from being occluded. The heavy braking could be a danger to traffic behind the AV in its lane.

## 5.2. Example 2: Traffic junction with VRU not following rules

In this recorded accident retrieved from Resembler's webtool [13], the vehicle with the dashcam recording is executing a right turn at a traffic junction with traffic light indicating a right turn green arrow. This was chosen as an example of other actors behaving against traffic rules creating complexity for an AV.

The traffic on the opposite direction has come to a stop. However, as the vehicle is executing its right turn with the right of way, an e-bike emerges from occlusion of the blue truck of the traffic in the opposing direction and continues straight against traffic lights and crashes into the recording vehicle. The driver attempts to stop upon first sighting of this e-bike emerging from occlusion but is unable to stop in time, resulting in a collision. From this accident, we will evaluate the complexity for an AV in the following analysis considering the AV as the camera vehicle. Vehicle 1 is the static truck, vehicle 2 is the e-bike in this case.

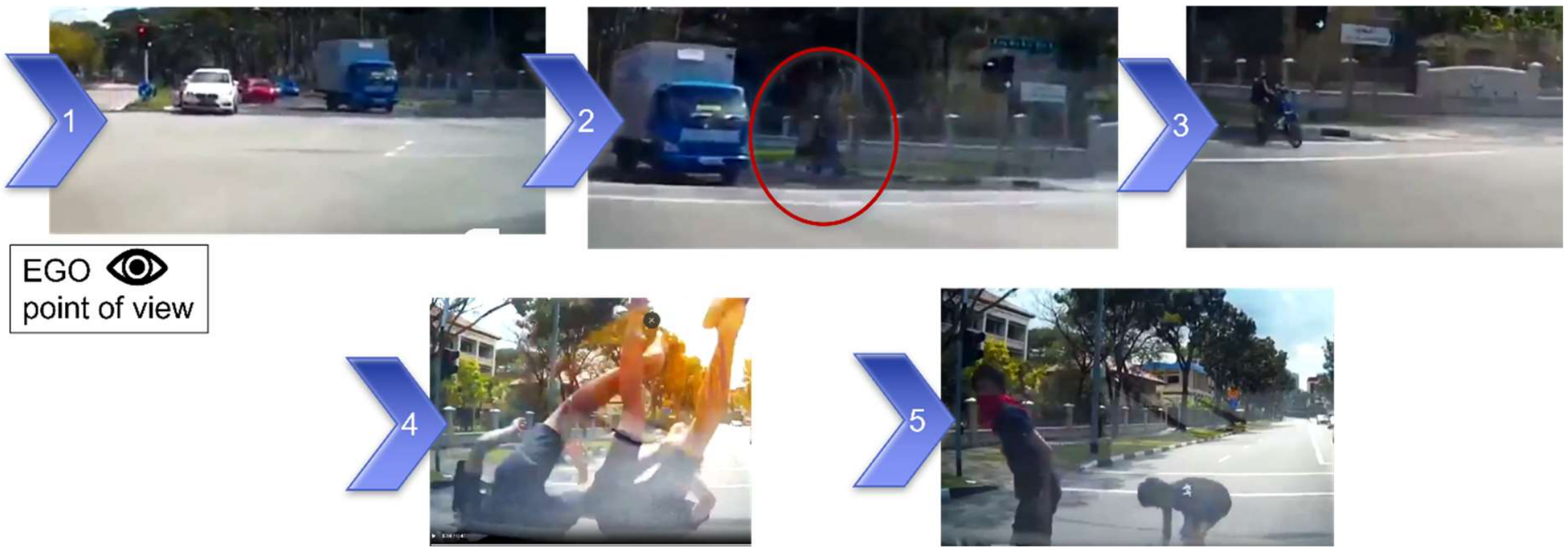

*Figure 4: Images from video recording of an accident where a car collides with an e-bike which had continued against red lights [13]*

### 5.2.1. Identify traffic agent hazard elements from guidelines and its associated complexity.

- **<u>Static Environment</u>**
  - Physical Infrastructure – Road infrastructure
    - Type- Traffic junction - Signalized
- **<u>Traffic Actor: Vehicle1 (truck)</u>**
  - Type- Nominal
  - State- Static
    - Nominal
- **<u>Traffic Actor: Vehicle2 (e-bike)</u>**
  - Type – Standard - e-bike
  - State -Dynamic
    - Visibility
    - Behaviour against traffic rules – Collision Point

*Table 2: Elements of traffic agent hazards in this accident and the associated complexity from the guidelines above.*

| **Elements of Traffic Agent Hazards** | | *Situation Awareness* | | | | Decision Making | |
|---|---|---|---|---|---|---|---|
| Traffic Agent Hazards | Details of traffic agent hazards | **Perception** | | **Situation Analysis** | | **Goal Management** | **Planning** |
| | | *-Position & State Estimation* | *-Detection & Tracking* | *-Context Awareness* | *-Prediction* | | |
| Static environment - Traffic junction - Signalized | Right of way indication | | | *AV may have difficulty understanding the pattern of traffic flow of other actors at intersection. It may not realise if others may be infringing rules of the road.* | *AV may have difficulty predicting the path of actors actions at these junctions.* | *AV may have difficulty in maintaining right of way versus safety distances to other vehicle.* | *AV may have difficulty in carrying out the execution manoeuvre into traffic maintaining safety distances to other vehicles.* |
| Actor: Vehicle2 – E-bike | Type- Nominal - standard e-bike | | *AVs might have difficulty in classifying typical vehicles on SG roads due to the variety such as trucks/lorries.* | *Limitation of the perception capability can lead to performance limitation such incorrect understanding of traffic flow well. E.g. E-bike users do not undergo traffic rules training as stringently as motorcyclists and may often flout traffic rules.* | *Limitation of the perception capability can lead to performance limitation such incorrect prediction of traffic flow* | *If AV has difficulty perceiving information of actors, it could affect its balancing of goals between traffic flow and safety to other actors.* | *Limitation of the perception capability can lead to performance limitation such as incorrect estimation of space available.* |
| | Dynamic State- Visibility | *AV is unable to have localisation information of occluded dynamic vehicle.* | *Unable to detect dynamic vehicle. No information for tracking of dynamic vehicles* | *If AV can't perceive a vehicle in its environment, it could lead to difficulty to understand the situation and to define the adapted AV behaviour.* | *If AV can't see a vehicle in its environment, it could lead to difficulty to predict the motion of this vehicle* | *Visibility of dynamic vehicle affects goal management. If the AV is not able to see the dynamic vehicle it will not be able to decide what is the action it should take in a safe way.* | *Visibility of dynamic vehicle affects knowledge of available space to carry out the planned manoeuvre.* |
| | Dynamic State-behaviour against traffic rules-Collision point-From side | | | *AV is not able to understand the traffic context and react accordingly to the situation.* | *Sudden and unusual behaviour of the actors are difficult to predict.* | *Each interaction with other illegal vehicles affects AV's goal of right of way versus safety to other actors.* | *Affects AV's available space since this is violated with actor's illegal manoeuvres.* |
| Actor: Vehicle1 - Truck | Type-Nominal | | *AVs might have difficulty in classifying typical vehicles on SG roads due to the variety such as trucks/lorries.* | *Limitation of the perception capability can lead to performance limitation such incorrect understanding of traffic flow well. E.g., big vehicles could be a source of occlusion.* | *Limitation of the perception capability can lead to performance limitation such incorrect prediction of traffic flow* | *If AV has difficulty perceiving information of actors, it could affect its balancing of goals between traffic flow and safety to other actors.* | *Limitation of the perception capability can lead to performance limitation such as incorrect estimation of space available.* |
| | Static State-Nominal-Lane allows for stopping of vehicles | | | *AV could have difficulty to understand the nominal interactions with other vehicles and react accordingly e.g. Awareness of parked vehicle (different from static vehicle in traffic)* | *AV with difficulty in context awareness may result in incorrect prediction of actor movement.* | *If AV has difficulty in context awareness, it could affect its balancing of goals between traffic flow and safety to other actors.* | *If AV has difficulty in context awareness, it can lead to performance limitation such as incorrect estimation of space available.* |

### 5.2.2. Overall analysis of each element in the context of the scenario

| Element | Element Category | Complexity Category | Complexity Reason |
|---|---|---|---|
| Static Environment-Traffic junction signalized | Right-of-way indication | **Context Awareness Prediction** | AV may have difficulty understanding the pattern of traffic flow of other actors at intersection. It may not realise if others are following or infringing rules of the road, and this could affect the prediction of other actor's actions at these junctions. |
| Vehicle (E-bike) | Dynamic - Visibility | **Detection and Tracking; Context Awareness** | If AV can't perceive the e-bike in its environment, it could lead to difficulty to understand the situation and to define the adapted AV behaviour. For a pattern recognition system, it takes some time to detect a e-cycle coming out of occlusion. Unlike the first example presented, here the actor occluded ( e-bike) was not visible to the EGO before entering in the occlusion, this has reduced the EGO's possibility to be aware about the situation. |
| | | **Prediction** | Difficulty to predict future trajectory of dynamic e-bike. |
| | | **Goal Management** | Occlusion of the moving e-bike affects an AV's goal management. If the AV is not able to see the dynamic vehicle it will not be able to decide what is the action it should take in a safe way, keeping in mind surrounding traffic. |
| Vehicle (E-bike) | Dynamic – Behaviour against traffic rule – Collision Point | **Prediction** | Sudden and unusual behaviours of the moving e-bike are difficult to predict. Especially at a controlled traffic intersection. |
| Vehicle (Truck) | Static | **Detection Context Awareness** | Large vehicle – AV might have difficulty in correct classification. It could also lead to occlusion of other actors. |
| | | **Context awareness** | Complexity of awareness of possible hidden actors behind. |

### 5.2.3. Final analysis of scenario complexity

This is an example of an accident where collision occurred with another road user that was behaving against traffic rules which is common scenario an AV could encounter on public roads. If the AV was approaching a traffic junction from the point of view of the dashcam vehicle, it is waiting traffic light to provide it right of way to execute a right turn. The traffic on the other side of the road such as the blue truck seen in Figure 4 comes to a stop while the AV is given a green right turn arrow. This truck as a traffic actor could be a source of complexity for the AV such as classification errors or context awareness that due to its size, it could occlude other actors. The AV would have understood from the lights and visible actors that it is given the right of way for the turn and its path is clear to start executing its move. This is the context awareness an AV should have of the infrastructure, understanding traffic flow and the corresponding traffic lights.

However, during the turn, a e-bike is seen emerging from behind the blue truck and continuing into the junction onto the collision path with the ego. This e-bike was occluded behind the truck and its detection is missed until the distance is very close to the AV. This is different from the earlier example where initial detection of pedestrian before occlusion allowed for the possibility of tracking and prediction, which is not present in this accident.

This occurrence of actors behaving against traffic rules is difficult for AVs, compounded with additional complexity element of the occlusion of the e-bike that could occur as seen in this accident event. If the AV detects this actor late, it has to suddenly manage its goals of maintaining safety distance to this other road user versus maintaining right of way and traffic flow behind. If it suddenly brakes to avoid collision and have safe distance to the offending e-bike, it could cause a collision from the traffic from behind. An AV could also attempt to turn further away from the e-bike but that could be a sudden lane change and an AV would need to check if the adjacent lane has sufficient space from other road users for a safe manoeuvre. In this instance of recording, the dashcam vehicle possibly could not brake in time and had a collision with the e-bike and its users. However thankfully the riders appeared to be able to recover and move to the side of the road. The actors listed in this analysis of complexity are those that directly interact and create complexity for an AV in this event, there could be other actors in the vicinity but are not considered in the analysis. It is important to note that on public roads and in this event, there were other actors behind the blue truck and those behind the dashcam vehicle that constitutes to normal traffic flow that could be around an AV and affects its goal management capability. Additionally, it is worthy to note that e-bikes and motorcycles are known to exhibit lane-splitting behaviour when travelling on roads, and this could be an additional complexity to the scenario. An AV could assume that all lanes for the oncoming have a vehicle in its lane which indicates that all traffic in the oncoming direction has stopped and could miss the possibility that e-bikes or motorcycles could be arriving at the traffic intersection between the vehicles in its lanes.

## 5.3. Example 3: Work zone and sudden lane change

This is an example of an accident where elements of traffic management zone as a traffic agent hazard is involved. In the recorded accident, temporary works are being carried out in the middle two road lanes out of four lanes on the road, with traffic flowing on the right and left lane of the work-zone. The vehicle with the dashcam recording is travelling on the right lane and captures the moment after the work zone where two vehicles ahead (white vehicle and yellow taxi) attempt to carry out abrupt lane change to reach the exit lane on the left shortly after the work zone. The abrupt lane change causes it to crash into a black vehicle on the most left lane. In this scenario, it will be analysed with an AV as the victim car on the left to capture the complexity an AV could need to handle when other actors could behave erratically on public roads, possible due to road works. Vehicle 1 in this analysis is the white vehicle, and the vehicle 2 is the yellow taxi.

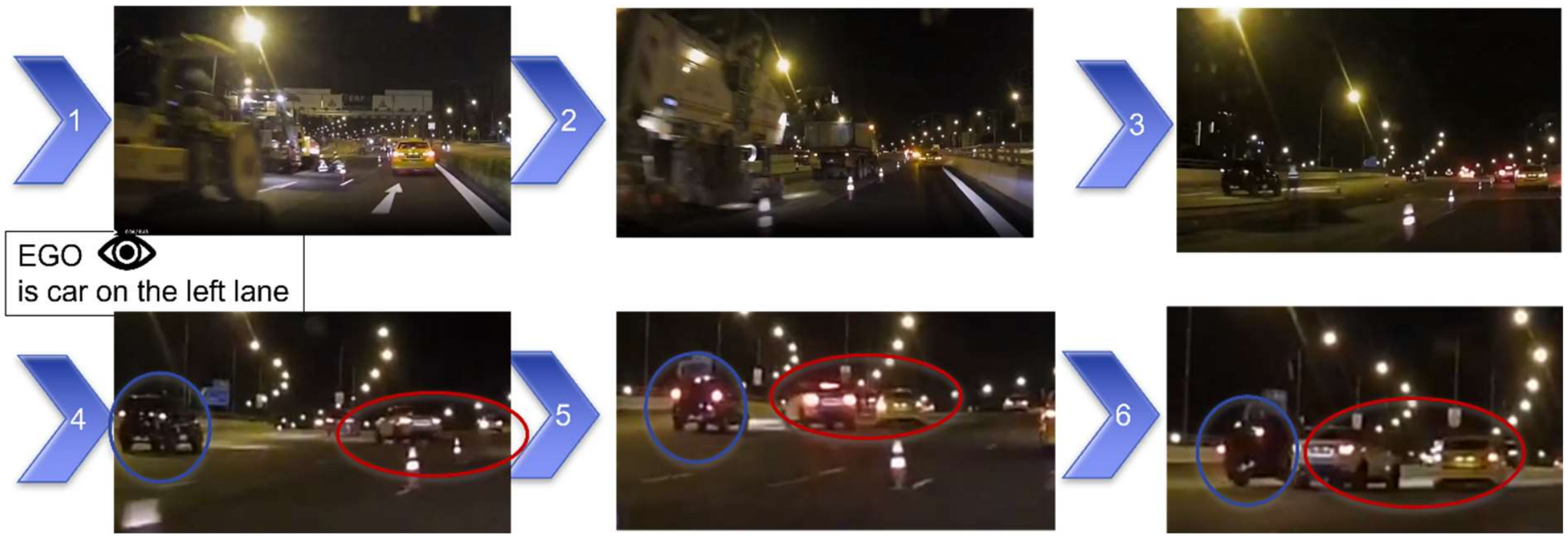

*Figure 5: Images from video recording of an accident where two cars collide after a work-zone area. [13]*

### 5.3.1. Identify traffic agent hazard elements from guidelines and its associated complexity.

- **Traffic Actor: Vehicle1 (white vehicle)**
  - Lights
  - State - Dynamic
    - Erratic – Lane Change
- **Traffic Actor: Vehicle 2 (yellow taxi)**
  - Lights
  - State - Dynamic
    - Visibility
    - Erratic – Lane Change
- **Traffic Management Zone**
  - Work-zone – (Elements)
- **Environmental Conditions**
  - Visibility – Illumination

*Table 3: Elements of traffic agent hazards in this accident and the associated complexity from the guidelines above.*

| **Elements of Traffic Agent Hazards** | | *Situation Awareness* | | | | Decision Making | |
|---|---|---|---|---|---|---|---|
| Traffic Agent Hazards | Details of traffic agent hazards | **Perception** | | **Situation Analysis** | | **Goal Management** | **Planning** |
| | | *-Position & State Estimation* | *-Detection & Tracking* | *-Context Awareness* | *-Prediction* | | |
| Actor: Vehicle 1 & 2 | Dynamic State-Erratic-Lane Change | | | *AV could have difficulty understand the context of actor's intention.* | *Sudden and unusual behaviours of the Actors (here moving vehicle) are difficult to predict.* | *Each interaction with other erratic vehicles affects AV's goal of driving in lane and sudden need to maintain safe distance and traffic flow.* | *Available space for manoeuvre to react to actor's erratic movement.* |
| Actor: Vehicle 1 & 2 | Vehicle Lights | | *AV might have difficulty in classifying the state of lights on other vehicles.* | *If AV has difficulty in getting state of actor vehicle's lights, it could lead to difficulty to understand the situation and to define the adapted AV behaviour.* | *If AV has difficulty in getting state of actor vehicle's lights, it could lead to difficulty to predict actor vehicle's movement.* | *If the AV is not able to understand the actor's intention, it will not be able to decide what is the action it should take in a safe way. i.e., understanding traffic flow versus right of way.* | *AV might have difficulty carrying out the planned manoeuvre due to wrong space estimation.* |
| Actor: Vehicle 2 | Dynamic State – Visibility-Other users occluding actor (vehicle 2) | *AV is unable to have localisation information of occluded dynamic vehicle.* | *Unable to detect dynamic vehicle. No information for tracking of dynamic vehicles* | *If AV can't perceive a vehicle in its environment, it could lead to difficulty to understand the situation and to define the adapted AV behaviour.* | *If AV can't see a vehicle in its environment, it could lead to difficulty to predict the motion of this vehicle* | *Visibility of dynamic vehicle affects goal management. If the AV is not able to see the dynamic vehicle it will not be able to decide what is the action it should take in a safe way.* | *Visibility of dynamic vehicle affects knowledge of available space to carry out the planned manoeuvre.* |
| Traffic Management Zone | Work-Elements - Cones | | *Where work zone elements are present, the AV might have difficulty in correct classification.* | *Difficulty in understanding the meaning of the temporary signs* | *If AV has difficulty in understanding work-zone elements,* | *Could be difficult to achieve AV's target if lane is closed (changing* | *AV could have difficulty in accurate estimation space* |
| | | | *The performance of the classification will depend on the previous training data available. Difficulty is linked to the diversity of the configurations involved.* | *associated to the zones and define a good behaviour for each situation encountered (e.g slow down, stop, works ahead)* | *it will not be able to predict the behaviour of the other road users who are able to perceive correctly the sign and could have a different behaviour than expected.* | *lane /rerouting).* | *availabl e for its planned manoeu vres.* |
| Environmenta l Conditions | Visibility – Illuminatio n- Low levels | *Limitation of the perception capability can lead to performanc e limitation such incorrect measureme nts* | *Limitation of the perception capability can lead to performance limitation such incorrect classification, incorrect tracking, misdetection. The system might not be able to detect objects with the same accuracy as it would be in more favorable conditions. E.g in low light conditions, objects might not reflect enough light, leading to reduced detection accuracy for a Lidar. Cameras might not be able to capture enough light to create a detailed accurate image.* | *Limitation of the perception capability can lead to an AV's incorrect understandin g of traffic flow as they might not be able to detect accurately obstacles, pedestrian, or other vehicles in their path.* | *Limitation of the perception capability can lead to incorrect prediction of traffic flow* | *This affects goal management of driving safely in the ODD. The AV may have difficulty in evaluation of the severity of visibility hazards* | *Limitation of the perception capability can lead to performan ce limitation such as incorrect estimation of space available.* |

### 5.3.2. Overall analysis of each element in the context of the scenario

| Element | Element Category | Complexity Category | Complexity Reason |
|---|---|---|---|
| Vehicle1 and 2 | Dynamic - Erratic - Lane Change | **Prediction**<br>**Goal Management** | Sudden and unusual behaviour of the Actors (here moving vehicle) are difficult to predict.<br>This affects an AV sudden need to maintain safe distance to these vehicles cutting into ego's lane from two lanes away and traffic flow around in case harsh braking is needed or required to change lane. |
| Vehicle1 and 2 | Vehicle Lights | **Detection and Tracking**<br>**Prediction** | AVs might have difficulty in classifying the state of light indicators from the 2 actor vehicles intention to change lanes – from two lanes away. This could lead to difficulty to understanding the 2 actor vehicles' intention and prediction of their path. |
| Vehicle2 | Dynamic - Visibility | **Detection and Tracking;**<br>**Context Awareness** | Complexity of detection with occlusion as it is behind vehicle 1. Hence unable for an AV to be aware of vehicle 2's situation. In this case, vehicle 2's action could have influenced vehicle 1's movement. |
| | | **Prediction** | Difficulty to predict future trajectory for this vehicle 2. |
| | | **Goal Management** | Occlusion of the moving this vehicle 2 affects an AV's goal management of safety to other traffic actors and its own right of way. If the AV is not able to see the dynamic vehicle it will not be able to decide what is the action it should take in a safe way |
| Traffic Management Zone | Work zone Elements (Cones indicating Lane Closure) | **Context Awareness**<br>**Goal Management** | Could be difficult to achieve actor's target when multiple lanes are closed due to work-zone elements. An AV might not have the awareness that the other actors (vehicles 1&2) were not able to change lanes due to the lane closures (indicated by work zone elements such as cones) which could have led to erratic lane change behaviour at the end of the work zone.<br>If an AV was aware – it could attempt to manage safety distances and speed in anticipation of such actor's movement. |
| Environment Conditions | Visibility -Illumination – Low levels (Nighttime) | **Goal Management** | This affects goal management of driving safely in the ODD. The AV may have difficulty in evaluation of the severity of visibility such as good detection of a car at night (vehicle 1) and with reduced perception could lead to reduced levels of prediction and context awareness of actors. |

### 5.3.3. Final analysis of scenario complexity

This was an event of an accident that resulted in a collision. In this instance it was evaluated from the point of view where an AV (ego) could be the vehicle on the most left (black vehicle). The event occurred at night, where there was work zone of lane closure occurring over a long stretch on the highway. This could also happen on non-highway road sections. Near the collision location, it could be seen that were was a highway exit to the left of the AV. For a different human driver, some might have anticipated that the long stretch of road closure could result in some other vehicles attempting to make multiple lane changes at the end of the work zone to reach their target goal of the exit lane. With this anticipation, the vehicle on the left might manage its speed or have a greater lookout area of where another vehicle might be crossing multiple lanes instead of just looking out for vehicles in the adjacent lane.

To safely operate through this traffic situation, the AV must detect roadworks in a reliable way, it could be but not exhaustive to these two following methods. 1. from the map (that requires a real-time ODD monitoring to update HD map) or 2. by recognizing work-zone elements (that requires a high perception capability to detect signs, cones, barriers). SAE Level 4 and 5 vehicles must have this capability for detecting roadworks. If roadworks is not part of the ODD of the AV with Level 4, the vehicle must disengage and perform a safe stop after the detection of the roadworks.

The AV is expected to recognize the intention of the actor vehicles driving aggressively and predict the possible cutting across the lanes (or double lane cut) to adapt its behavior. This type of assessment should only be done by simulation for safety reasons. In addition, the AV perception system is expected to detect the lights indicators of the actors to increase the situation awareness and avoid the accident.

Some AVs might not have the capability for this context awareness and adapted behaviour of extended detection area to adjust to actors making an erratic lane change from multiple lanes away. This is compounded with the night environmental conditions where illumination is lower. This could add on to complexity for AVs where detection capability is reduced at night. This accident had two vehicles attempting to have sudden lane change, and the vehicle 2 in yellow (ahead of the vehicle 1 in white) could have affected the vehicle 1's actions due to its space available. An AV might not have detected this second yellow vehicle due to occlusion as well.

Lane closures from temporary work-zones, compounded with night conditions could lead to other road users to have erratic behaviour and this accident is evident of one situation that would be difficult for an AV to operate in due to multiple elements of complexity. Management of temporary work-zones by authorities should aim to reduce possibilities of human driving erratic behaviour in future interactions with AVs on road, such as the late change for an exit as seen in this accident.

## 5.4. Example 4: Reachability with traffic flow

This accident was chosen to highlight the complexity for an AV when its reachability is difficult to achieve. In this accident, the blue vehicle ahead of the vehicle with the dashcam recording is shown be travelling on the left lane of a dual lane road and appears to be wanting to reach the right filter lane but could not do so initially due to traffic in the adjacent lane like the yellow vehicle on the right. However, as the blue vehicle makes an abrupt lane change just before the right filter lane but did not notice a motorcycle travelling in the adjacent lane (possibly in its blind spot) and collides with this motorcycle.

The complexity for an AV would be if the blue vehicle was an AV travelling on a road attempting to reach a right filter lane, but this reachability may not be achieved due to traffic in the adjacent lane such as the yellow taxi as vehicle 1 and motorcycle as vehicle 2. The camera vehicle could an actor as vehicle 3 to represent the traffic flow behind the AV.

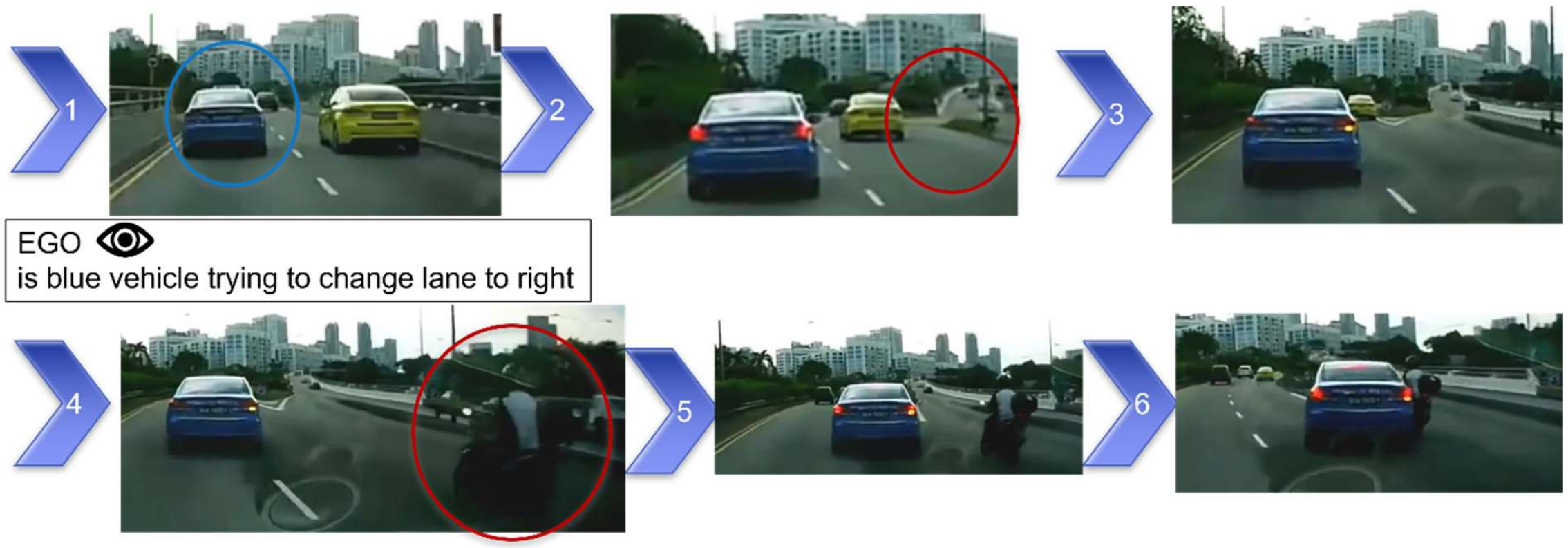

*Figure 6: Accident scenario of a blue vehicle colliding into a motorcycle while trying to change lane.*

### 5.4.1. Identify traffic agent hazard elements from guidelines and its associated complexity.

- **<u>Traffic Actor: Vehicle1 (yellow taxi)</u>**
  - State - Dynamic
    - Nominal - Parallel
- **<u>Traffic Actor: Vehicle 2 (motorcycle)</u>**
  - Type-Nominal- Motorcycle
  - State - Dynamic
    - Visibility
- **<u>Traffic Actor: Vehicle 3 (camera vehicle)</u>**
  - State- Dynamic
    - Nominal - Following (actor behind ego)

*Table 4: Elements of traffic agent hazards in this accident and the associated complexity from the guidelines above.*

| **Elements of Traffic Agent Hazards** | | *Situation Awareness* | | | | Decision Making | |
|---|---|---|---|---|---|---|---|
| Traffic Agent Hazards | Details of traffic agent hazards | **Perception** | | **Situation Analysis** | | **Goal Management** | **Planning** |
| | | *-Position & State Estimation* | *-Detection & Tracking* | *-Context Awareness* | *-Prediction* | | |
| Actor: Vehicle 1 | Dynamic State– Nominal - Parallel | | | *AV could have difficulty understand the nominal interactions with other vehicles and react accordingly.* | *AV with difficulty in context awareness may result in incorrect prediction of actor movement.* | *AV could have difficulty balancing of goals between maintaining right of way and traffic flow and safety to other actors. E.g., Managing safe lateral clearance should not be compromised, slow down and stay behind vehicle or speed up.* | *If AV has difficulty in context awareness. It can lead to performance limitation such as incorrect estimation of space available.* |
| Actor: Vehicle 2 | Type – Nominal- Standard- Motorcycl e | | *AVs might have difficulty in classifying typical vehicles on SG roads due to the variety such as trucks/lorries .* | *Limitation of the perception capability can lead to performance limitation such incorrect understanding of traffic flow well.* | *Limitation of the perception capability can lead to performance limitation such incorrect prediction of traffic flow* | *If AV has difficulty perceiving information of actors, it could affect its balancing of goals between traffic flow and safety to other actors.* | *Limitation of the perception capability can lead to performance limitation such incorrect estimation of space available.* |
| | Dynamic State– Visibility- Other users occluding actor (vehicle 2) | *AV is unable to have localisation informatio n of occluded static vehicle.* | *Unable to detect dynamic vehicle. No information for tracking of dynamic vehicles* | *If AV can't perceive a vehicle in its environment, it could lead to difficulty to understand the situation and to define the adapted AV behaviour.* | *If AV can't see a vehicle in its environment, it could lead to difficulty to predict the motion of this vehicle* | *Visibility of dynamic vehicle affects goal management. If the AV is not able to see the dynamic vehicle it will not be able to decide what is the action it should take in a safe way.* | *Visibility of dynamic vehicle affects knowledge of available space to carry out the planned manoeuvre.* |
| Actor: Vehicle 3 | Dynamic State – Nominal (actor following ego) | | | | | *AV could have difficulty managing traffic flow behind it when carrying out other goals from interaction with other actors.* | *AV could have difficulty in deciding manoeuvre to manage the space available to actor behind.* |

### 5.4.2. Overall analysis of each element in the context of the scenario (event)

| Element | Element Category | Complexity Category | Complexity Reason |
|---|---|---|---|
| Vehicle1 | Dynamic - Nominal - Parallel | **Goal Management** | Reachability vs safety to other road user's vs traffic flow. In this instance the AV is waiting for available space to move into adjacent lane |
| Vehicle2 | Type- Nominal - Motorcycle | **Detection and Tracking** | AVs might have difficulty in classifying typical vehicles on SG roads due to the variety on the roads. Some AVs might have difficulty detecting a motorcycle due to the smaller surface area. |
| | Dynamic - Visibility occlusion | **Detection and Tracking:** | If AV can't see the motorcycle in its environment due to it emerging from behind other cars in the adjacent lane, it could lead to difficulty to understand the situation and to define the adapted AV behaviour. |
| | | **Prediction** | Difficulty to predict behaviour of possibly occluded motorcycle. |
| | | **Goal Management** | Visibility of dynamic vehicle affects goal management.<br>If the AV is not able to see the dynamic vehicle it will not be able to decide what is the action it should take in a safe way. |
| | Dynamic - Nominal-Parallel | **Goal Management** | Reachability vs safety to this road user (motorcycle) |
| Vehicle3 | - Dynamic - Nominal - Following (actor following ego) | **Goal Management** | Safety to other road user's vs traffic flow. If there is no traffic behind AV such as vehicle 3, AV might not have issue of managing traffic flow, and could wait in the lane for an opportunity to change lane and achieve its reachability goal, |

### 5.4.3. Final analysis of complexity of scenario

In this incident, an accident occurred between the blue vehicle and the motorcycle in the adjacent lane when the blue vehicle executes a late lane change to attempt to go into a slip road on the right side of the road. This could be a complex scenario for an AV if it was in the traffic situation like the blue vehicle. An AV would typically keep left of a two-lane carriageway unless overtaking [15]. In this scenario, an AV needs to filter to the right lane in order to reach its target lane, however it is keeping to the left of the two-lane carriageway and is looking for an opportunity to carry out a lane change into the right lane. It could let traffic in the adjacent lane pass such as the yellow vehicle 1 in Figure 6 above. This could happen multiple times in the lead up to approaching the slip road, however when the AV is near the slip road, it could however have poor goal management between its reachability against managing safety to other road users such as the motorcycle. This complexity in managing an AV's goal management in a busy road traffic environment could occur on Singapore's public roads and may occur outside a highway setting with an AV needing to carry out a right turn at a junction ahead and a lane change is required to enter a right turning lane.

## 5.5. Example 5: Sudden cut out and sudden braking

This is an accident that occurred due to erratic actor behaviour. From the point of view of the vehicle with the dashcam travelling on the right of three lanes, it is travelling behind a white vehicle, with traffic flowing on the middle lane (adjacent of the recording vehicle) and sometimes with a motorcycle that is lane splitting. It was seen the white vehicle ahead suddenly brakes and does an abrupt left lane change, colliding with the lane splitting motorcycle that was probably in the blind spot of the white vehicle. It can be seen the black vehicle right ahead of the white vehicle had come to an abrupt stop, and the recording vehicle barely manages to brake and stop just before possible collision with this black vehicle. This could be a complex scenario for an AV as the car with the dashcam, and other actors surrounding it are the black vehicle ahead as vehicle 1, the white vehicle ahead as vehicle 2 and the motorcycle in the adjacent lane as vehicle 3.

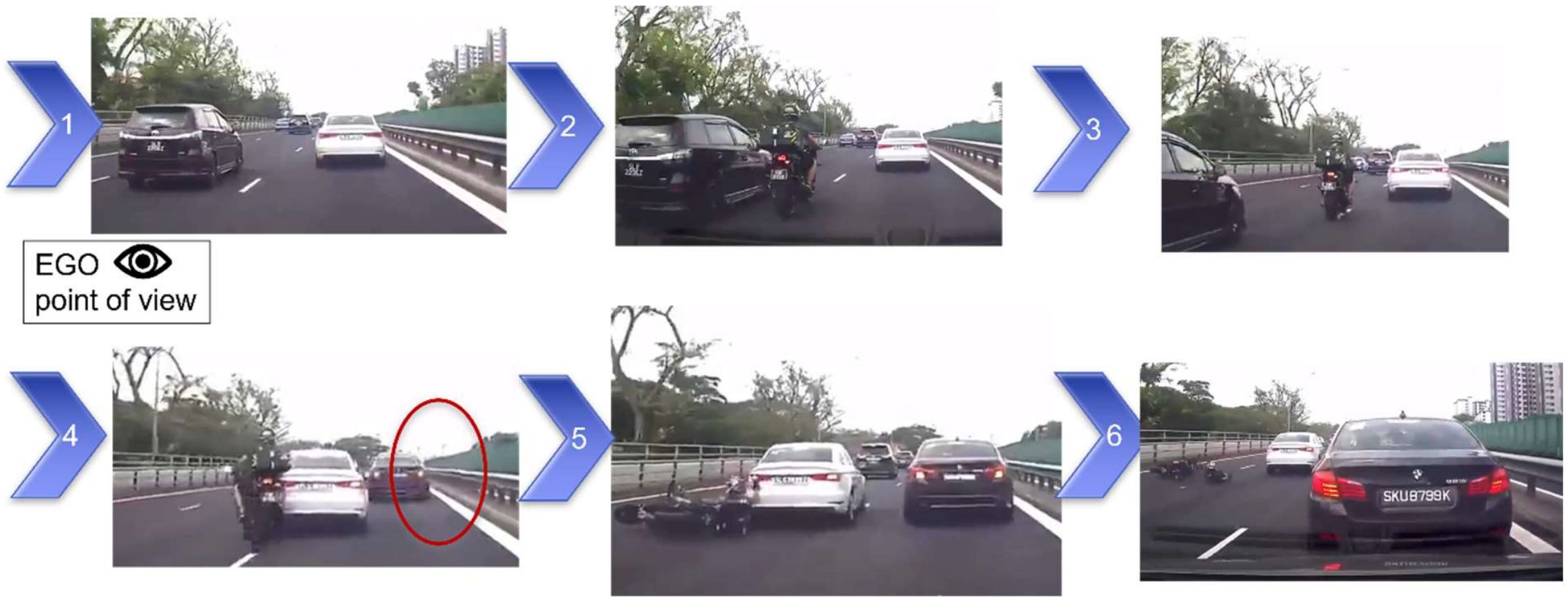

*Figure 7: Accident event where vehicle ahead suddenly cuts out.*

### 5.5.1. Identify traffic agent hazard elements from guidelines and its associated complexity.

The AV would be in the point of view of the vehicle with the recording dashcam.

- **<u>Traffic Actor: Vehicle 1 (Black vehicle)</u>**
  - State - Dynamic
    - Visibility
    - Erratic – Following (Sudden braking)
- **<u>Traffic Actor: Vehicle2 (White vehicle)</u>**
  - State - Dynamic
    - Erratic – Following (Sudden cut out)
- **<u>Traffic Actor: Vehicle 3 (Motorcycle)</u>**
  - Type -Nominal - (Motorcycle)
  - State - Dynamic
    - Erratic – Parallel (Lane splitting)

*Table 5: Elements of traffic agent hazards in this accident and the associated complexity from the guidelines above.*

| **Elements of Traffic Agent Hazards** | | *Situation Awareness* | | | | Decision Making | |
|---|---|---|---|---|---|---|---|
| Traffic Agent Hazards | Details of traffic agent hazards | **Perception** | | **Situation Analysis** | | **Goal Management** | **Planning** |
| | | *-Position & State Estimation* | *-Detection & Tracking* | *-Context Awareness* | *-Prediction* | | |
| Actor Vehicle 1 (Black) | Dynamic State-visibility-Other users occluding actor (vehicle 1) | *AV is unable to have localisation information of occluded dynamic vehicle.* | *Unable to detect dynamic vehicle. No information for tracking of dynamic vehicles* | *If AV can't perceive a vehicle in its environment, it could lead to difficulty to understand the situation and to define the adapted AV behaviour.* | *If AV can't see a vehicle in its environment, it could lead to difficulty to predict the motion of this vehicle* | *Visibility of dynamic vehicle affects goal management. If the AV is not able to see the dynamic vehicle it will not be able to decide what is the action it should take in a safe way.* | *Visibility of dynamic vehicle affects knowledge of available space to carry out the planned manoeuvre.* |
| | Dynamic State-Erratic – Following ( Sudden braking) | | | *AV could have difficulty understand the context of actor's intention.* | *Sudden and unusual behaviours of the Actors are difficult to predict.* | *Each interaction with other erratic vehicles affects AV's goal of driving in lane and sudden need to maintain safe distance and traffic flow.* | *AV could have difficulty in estimating available space for manoeuvre to react to actor's erratic movement.* |
| Actor Vehicle2 (White) | Dynamic State-Erratic – Following (sudden cut out) | | | *AV could have difficulty understand the context of actor's intention.* | *Sudden and unusual behaviours of the Actors are difficult to predict.* | *Each interaction with other erratic vehicles affects AV's goal of driving in lane and sudden need to maintain safe distance and traffic flow.* | *AV could have difficulty in estimating available space for manoeuvre to react to actor's erratic movement.* |
| Actor Vehicle 3 | Type-Nominal-Standard (Motorcycle) | | *AVs might have difficulty in classifying typical vehicles on SG roads due to the variety such as trucks/lorries.* | *Limitation of the perception capability can lead to performance limitation such incorrect understanding of traffic flow well.* | *Limitation of the perception capability can lead to performance limitation such incorrect prediction of traffic flow* | *If AV has difficulty perceiving information of actors, it could affect its balancing of goals between traffic flow and safety to other actors.* | *Limitation of the perception capability can lead to performance limitation such incorrect estimation of space available.* |
| | Dynamic State - Erratic – Parallel- | | | *AV could have difficulty understand the context of* | *Sudden and unusual behaviours of the Actors* | *Each interaction with other erratic vehicles affects AV's goal* | *Available space for manoeuvre to react to* |
| | Actor not keeping in lane (Lane splitting) | | | *actor's intention. E.g., motorcycle that is lane splitting might suddenly cut into lane.* | *are difficult to predict.* | *of driving in lane and sudden need to maintain safe distance and traffic flow.* | *actor's erratic movement.* |

### 5.5.2. Overall analysis of each element in the context of the scenario (event)

| Element | Element Category | Complexity Category | Complexity Reason |
|---|---|---|---|
| Actor Vehicle 1 (Black) | Dynamic- Visibility | **Detection and Tracking; Prediction** | If AV can't see this vehicle1 (black car) in its environment it could lead to difficulty for an AV to understand the situation and to define the adapted AV behaviour, leading to difficulty in prediction of behaviour |
| | | **Goal Management; Planning** | Visibility of this dynamic vehicle1 affects an AV's goal management.<br>If the AV is not able to see the dynamic vehicle it will not be able to decide what is the action it should take in a safe way and the available space for manoeuvre. |
| | Dynamic -Erratic – Following (Sudden braking) | **Prediction** | Sudden and unusual behaviour of the Actors (here moving vehicle) are difficult to predict. |
| Actor Vehicle2 (White) | Dynamic - Erratic – Following (sudden cut out) | **Prediction** | Sudden and unusual behaviour of the actor vehicle 2 are difficult to predict. |
| | | **Context Awareness** | After sudden movement of vehicle cutting out, would the AV have awareness that this vehicle2 is cutting out due to sudden braking of the vehicle right ahead of it (vehicle1). |
| Actor Vehicle3 | Type (Motorcycle) | **Detection** | AVs might have difficulty in classifying typical vehicles on SG roads due to the variety on the roads. Some AVs might have difficulty detecting a motorcycle due to the smaller surface area. |
| | Dynamic -Erratic – Parallel | **Detection and Tracking:** | If AV can't see this motorcycle from behind (lane splitting) in its environment it could lead to difficulty to understand the situation and to define the adapted AV behaviour. |
| | | **Goal Management** | Safety to other road vehicles (vehicle 1 braking and vehicle3 motorcycle) vs traffic flow. Should AV brake or change lane? |

### 5.5.3. Final analysis of complexity of scenario

In this accident event, it could be a complex scenario where the ego vehicle could be the vehicle with the dashcam recording following a white vehicle ahead. This white vehicle does a sudden lane change to its left and did not see a motorcycle that was riding in between lanes on its left, resulting in a crash between the motorcycle and the white vehicle. After the white vehicle sudden lane change (cutting out of lane), it is seen that the black vehicle ahead of the white vehicle had come to a sudden stop. This results in the dashcam vehicle having to brake hard. This accident could be a complex scenario for an AV when the vehicle ahead suddenly cuts out of lane, with the AV unaware of the traffic flow ahead that was previously occluded by the vehicle ahead such as the white car. An AV would need to react to sudden actions of actors and determine quickly what is the best action to execute safely next, either to brake hard or carry out a sudden lane change.

A motorcycle as a traffic actor brings additional complexity to the AV. Motorcycles could be a source of complexity due to its size and dynamic behaviour which the AV should be able to detect and to consider its common characteristic “lane splitting” behaviour, allowing the AV to be able to react accordingly to the situation. AVs would need to be aware that a motorcycle could pass and overtake the AV anytime and be capable of providing sufficient lateral space when it occurs.

In this example, a motorcycle could be alongside the AV at this point of decision making and this would have to be considered in the AV’s goal management and consequently its planning of the appropriate manoeuvre. The complexity elements of predicting erratic actor behaviour and the balancing of goals in this situation make this accident an example of a scenario with multi agents with its corresponding complexity elements that would help to assess the AV.

# 6. Conclusion

To support the testing of AVs today beyond simplistic scenarios or edge cases where parameters are setup to the extreme, CETRAN has created a guideline for the evaluation of complex multi agent test scenarios. This allows for a clear structured manner in evaluating complexity elements incorporated in test scenarios based on the corresponding difficulties an AV might encounter in Singapore traffic.

The approach adopted in this study was to first understand the architecture of an AV system to enable insights on challenges an AV might experience on road. The next step was to consolidate complexity sources for AVs from traffic agent hazards with this understanding. A list of these traffic agent hazard and corresponding classification and examples are shown in Section 4.2 above. One complexity element introduced in a test scenario could also create a complex scenario for assessment of an AV, without requiring multiple complexity elements to be present. Some of the complexity from traffic agent hazards rely on the presence of other actors in the scenario such as pedestrians occluded by other vehicle, which leads to multi-agent complex scenarios.

These guidelines could be used for advancing current test scenarios through the following ways:

1. Adding complexity to the current test scenarios by incorporating new elements from the structured database (e.g. weather conditions impacting the visibility, road condition limiting the AV adherence)
2. Create new test scenarios targeted to assess different functionalities of AVs capabilities (perception and situation analysis within situation awareness versus decision making focused scenarios). This could integrate a certain level of granularity per example considering elements to the scenarios focused on the difficulties of perception as classification (e.g., pedestrian with accessories) or tracking (e.g., actor occluded temporarily). Scenarios could also be focused on the assessment of AV's capability to be aware of the surrounding context (e.g., different interactions with actors, interference zones, special zones as school zone, different types of infrastructures such as junctions, roundabouts, or merging lanes). Another focus could be on decision making component, adding elements or group of elements that challenge the goal management strategy of the AV (e.g., Multiple traffic agent hazard elements could be used for testing advanced AV capabilities for managing multiple goals such as safety, traffic flow, reachability)

Incorporating elements of complexity to modify or create new scenarios could be varied through the parameters set and result in several variations impacting the risk and severity of the test. It depends on the need of the scenarios that is created and the outcome it aims to achieve. This could lead to chain effect scenarios where the complexity of goal management becomes more severe based on parameters used for triggering actor interactions.

These guidelines were also applied for the analysis of real-traffic situations that were obtained from Resembler's database of on-road accidents and incidents in Singapore. This demonstrated the practicality of the guidelines created, implementing it in the analysis of real traffic situations with complex elements as a basis for scenarios that could be used to assess AVs.

# 7. Acknowledgements

We acknowledge our current partner who supported us in this project. We thank the team from Resembler for their valuable support in sharing insights on AV difficulties based on situations from their accident database.

This research/project is supported by the National Research Foundation, Singapore, and Land Transport Authority (Urban Mobility Grand Challenge (UMGC-L010)). Any opinions, findings and conclusions or recommendations expressed in this material are those of the author(s) and do not reflect the views of National Research Foundation, Singapore, and Land Transport Authority.

# 8. Acronyms

The terms and abbreviations used in the document are listed below.

| | |
|---|---|
| ADS | Autonomous Driving System |
| AI | Artificial Intelligence |
| AV | Autonomous Vehicle |
| CETRAN | Centre of Excellence for Testing & Research of Autonomous Vehicle |
| DGITM | General Directorate of Infrastructure, Transport and the Sea (France) |
| EMAS | Expressway Monitoring and Advisory System |
| FuSa | Functional Safety Standard |
| ISO | International Organization for Standardization |
| IRT | Institute for Technological Research |
| LTA | Land Transport Authority |
| ODD | Operational Design Domain |
| PAS | Publicly Available Specification |
| PMD | Personal Mobility Device |
| RCA | Research Collaboration Agreement |
| SAE | Society of Automotive Engineers (United States) |
| SOTIF | Safety of The Intended Functionality |
| VRU | Vulnerable Road User |

# 10. Appendices

## 10.1. Scenarios for Simulation

Four scenarios were designed for virtual creation using Carla (an open-source simulator) by the CETRAN team, applying the guidelines for complexity elements created in this work. This was used to illustrate the difficulties an ADS (Apollo in simulation) could experience with such scenarios. The four scenarios designed covers the variety of traffic agent hazard elements including common traffic actors of pedestrians, cars, and emergency vehicles, as well as work zones and common pedestrians jaywalking. Interaction with these actors at various common road intersections are also included to cover the various complex elements of scenarios from traffic situations in Singapore. The elements of traffic agent hazards in each scenario and the corresponding complexity reasons are tabulated below.

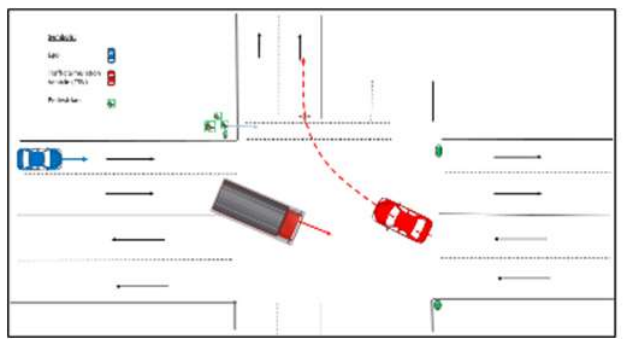
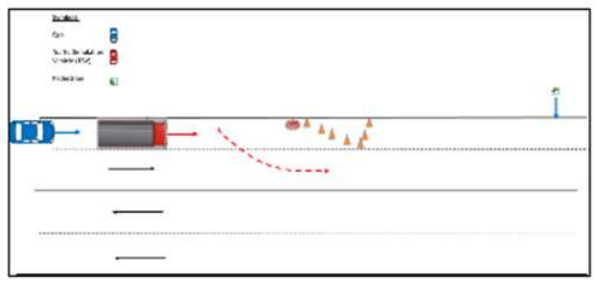
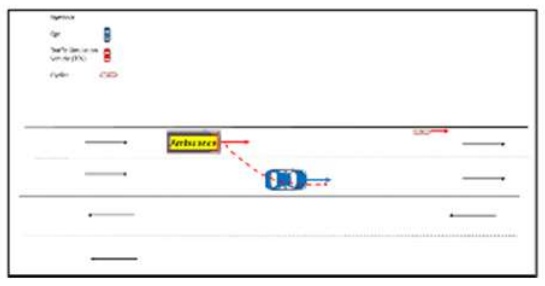
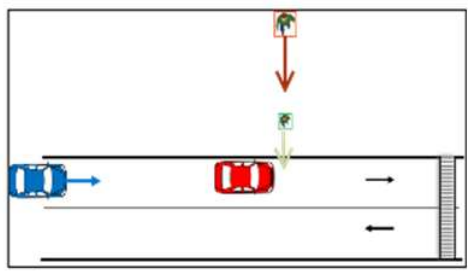

| **Scenario** | **- Traffic Intersection** | **- Work- zone** | **- Emergency Vehicle** | **- Stopped car** |
|---|---|---|---|---|
| **Elements of traffic agent hazards** | ○ **Actor: Vehicle**<br>• Static x2<br>• Visibility<br>○ **Actor: Pedestrian**<br>• Dynamic<br>• Static<br>○ **Actor: Object**<br>• Static – temporary<br>○ **Static Environment**<br>• Signalized pedestrian crossing<br>• Discretionary right turn | ○ **Actor: Vehicle**<br>• Dynamic<br>○ **Traffic Management Zone**<br>• Signs<br>• Work-zone element Lane closure<br>• Visibility<br>○ **Actor: Pedestrian**<br>• Dynamic | ○ **Actor: Vehicle**<br>• Static to Dynamic<br>• Special Type<br>○ **Actor: Cyclist**<br>• Dynamic | ○ **Actor: Vehicle**<br>• Static<br>○ **Actor: Pedestrian**<br>• Dynamic<br>• Visibility<br>• Type-Child<br>○ **Static Environment**<br>• Speed Hump<br>• Unsignalized - Pedestrian crossing |

| Scenario | - Traffic Intersection | - Work- zone | - Emergency Vehicle | - Stopped car |
|---|---|---|---|---|
| **Complexity for AVs** | ❖ **Situation Awareness**<br>- Possible occluded vehicle making discretionary right turn<br>-Classification object/passable or not?<br>-Awareness about pedestrian intention | ❖ **Situation Awareness**<br>- Classification of work-zone elements<br>-Awareness of lane closure ahead occluded behind vehicle. | ❖ **Situation Awareness**<br>- Classification of VRU, emergency vehicle<br>- Awareness of right of way of emergency vehicles | ❖ **Situation Awareness**<br>- Detect and track pedestrian<br>- Awareness of future pedestrian movement occluded behind vehicle.<br>- Awareness of parked vehicle vs static vehicle in traffic. |
| | ❖ **Decision making**<br>- Maintaining safety to occluded vehicle versus right of way of AV<br>- Drive over the object vs traffic flow (change lane/stop) | ❖ **Decision making**<br>-To change lane to maintain traffic flow versus ensuring safety when changing lane.<br>-Maintain safety to other road users vs right of way | ❖ **Decision making**<br>- Giving way to emergency vehicle vs safety to other road users. | ❖ **Decision making**<br>- Overtake or wait? Maintain traffic flow and safety distance. |